\documentclass[11pt, a4paper, logo, copyright, nonumbering]{main}
\usepackage[authoryear, sort&compress, round]{natbib}
\usepackage{dblfloatfix}
\usepackage{ulem}
\usepackage{caption}
\usepackage{dramatist}
\usepackage{xspace}
\usepackage{pifont} %
\usepackage{tcolorbox}
\tcbuselibrary{breakable} 
\usepackage{xltabular}
\usepackage{longtable}

\usepackage{amsfonts}
\usepackage{amsmath}
\usepackage{amssymb}
\usepackage{lineno}
\usepackage{multirow}
\usepackage{adjustbox}

\usepackage[bottom]{footmisc}

\usepackage{CJKutf8}
\usepackage{setspace}

\usepackage{dsfont}
\usepackage{array} %
\usepackage{subfigure} %
\usepackage{xcolor} %
\usepackage{tabularx}
\usepackage{booktabs}
\usepackage{xspace}

\usepackage{lipsum}  %
\usepackage{multicol} %

\usepackage{epigraph}
\usepackage{soul}
\usepackage{hyperref}
\usepackage{cleveref}

\usepackage{CJKutf8}

\makeatletter
\def\@BTrule[#1]{%
  \ifx\longtable\undefined
    \let\@BTswitch\@BTnormal
  \else\ifx\hline\LT@hline
    \nobreak
    \let\@BTswitch\@BLTrule
  \else
    \let\@BTswitch\@BTnormal
  \fi\fi
  \global\@thisrulewidth=#1\relax
  \ifnum\@thisruleclass=\tw@\vskip\@aboverulesep\else
  \ifnum\@lastruleclass=\z@\vskip\@aboverulesep\else
  \ifnum\@lastruleclass=\@ne\vskip\doublerulesep\fi\fi\fi
  \@BTswitch
}
\makeatother

\addto\extrasenglish{
}

{\begin{list}{}%
        {\setlength{\leftmargin}{#1}}%
        \item[]%
}
{\end{list}}

\reportnumber{001} %

\renewcommand{\today}{}

\title{\centering \modelname~Technical Report}

\def\huggingface{\raisebox{-1.5pt}{\includegraphics[height=1.05em]{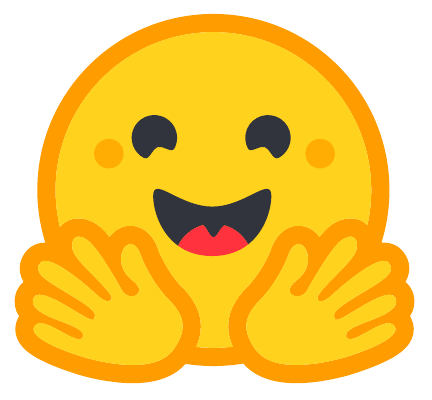}}}

\def\modelscope{\raisebox{-1.5pt}{\includegraphics[height=1.2em]{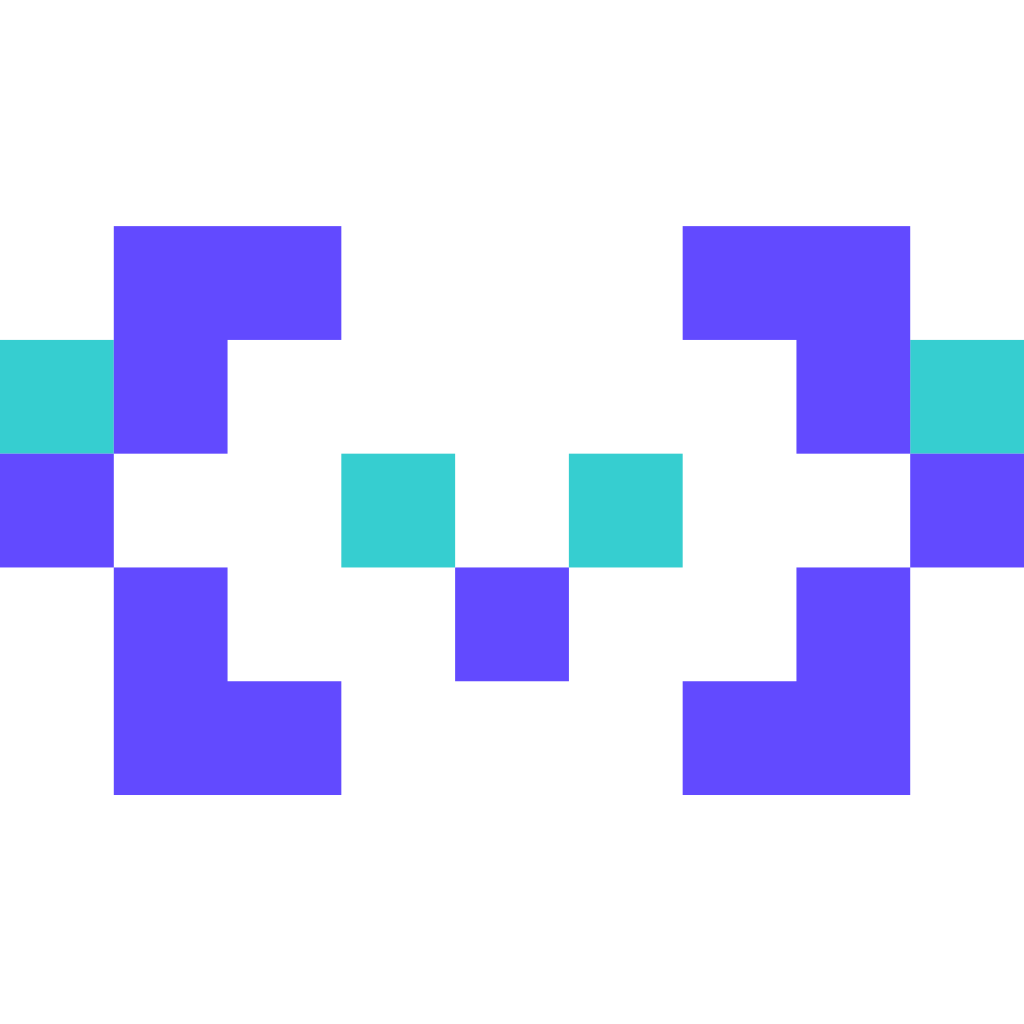}}}
\def\homepage{\raisebox{-1.5pt}{\includegraphics[height=1.2em]{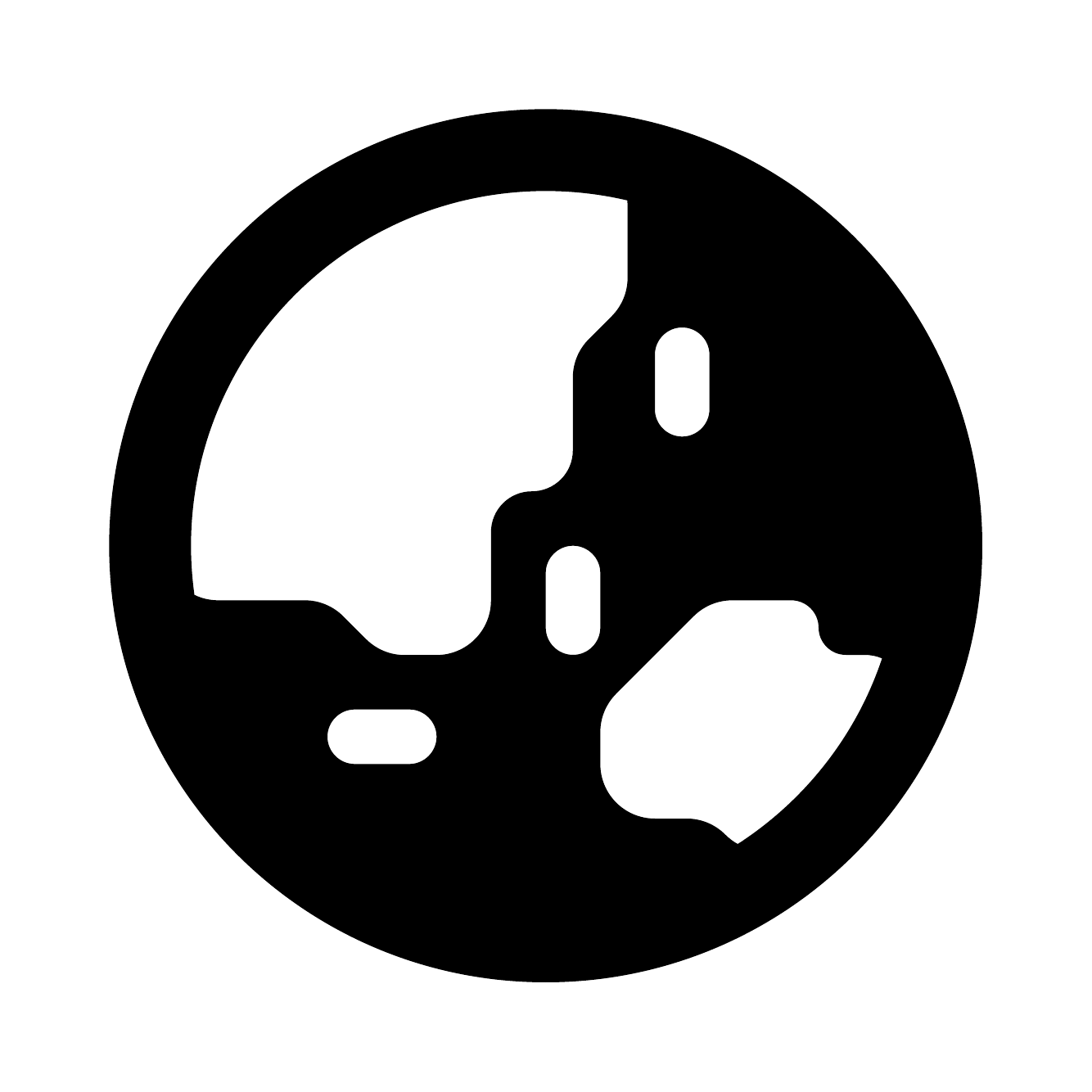}}}

\author[*]{
\textbf{Multimodal Intelligence Team, StepFun}
\\
\vspace{-0.7em}
\small \homepage~\textbf{Homepage}: \url{https://stepfun-ai.github.io/Step3-VL-10B} \\
\small \modelscope~\textbf{ModelScope}: \url{https://modelscope.cn/collections/stepfun-ai/Step3-VL-10B} \\
\small \huggingface~\textbf{Huggingface}: \url{https://huggingface.co/collections/stepfun-ai/step3-vl-10b}
}

\renewcommand{\phi}{\varphi}

\renewcommand{\epsilon}{\varepsilon}
\renewcommand{\imath}{\mathrm{i}}

\definecolor{ahared}{RGB}{209, 52, 56}

\newlength{\restsubwidth}
\newlength{\restsubheight}
\newlength{\restsubmoreheight}
\newcommand{\rest}[2]{%
        \settowidth{\restsubwidth}{\ensuremath{#2}}
        \settoheight{\restsubheight}{\ensuremath{{}_{#2}}}
        \ensuremath{{#1\hskip 0.5pt}_{\vrule\kern2pt\parbox[b][%
        4pt][b]{\the\restsubwidth}{%
                        \ensuremath{{}_{#2}}}}}
        }

\newcommand{\modelname}{\textsc{Step3-VL-10B}}

\newcommand{\promptbox}[1]{\noindent\fbox{\begin{minipage}{0.97\linewidth}\ttfamily #1\end{minipage}}}

\begin{abstract}

We present \textbf{\modelname}, a lightweight open-source foundation model designed to redefine the trade-off between compact efficiency and frontier-level multimodal intelligence. \modelname~is realized through two strategic shifts: first, a \textbf{unified, fully unfrozen pre-training strategy} on 1.2T multimodal tokens that integrates a language-aligned Perception Encoder with a Qwen3-8B decoder to establish intrinsic vision-language synergy; and second, a scaled post-training pipeline featuring \textbf{over 1k iterations of reinforcement learning}. Crucially, we implement Parallel Coordinated Reasoning (PaCoRe) to scale test-time compute, allocating resources to scalable perceptual reasoning that explores and synthesizes diverse visual hypotheses. Consequently, despite its compact 10B footprint, \modelname~rivals or surpasses models 10$\times$--20$\times$ larger (e.g., GLM-4.6V-106B, Qwen3-VL-235B) and top-tier proprietary flagships like Gemini 2.5 Pro and Seed-1.5-VL. Delivering best-in-class performance, it records 92.2\% on MMBench and 80.11\% on MMMU, while excelling in complex reasoning with 94.43\% on AIME2025 and 75.95\% on MathVision. We release the full model suite to provide the community with a powerful, efficient baseline.

\end{abstract}

\begin{document}

\maketitle

\vspace{1em}
\begin{figure}[h!]
\centering
\includegraphics[width=0.95\textwidth]{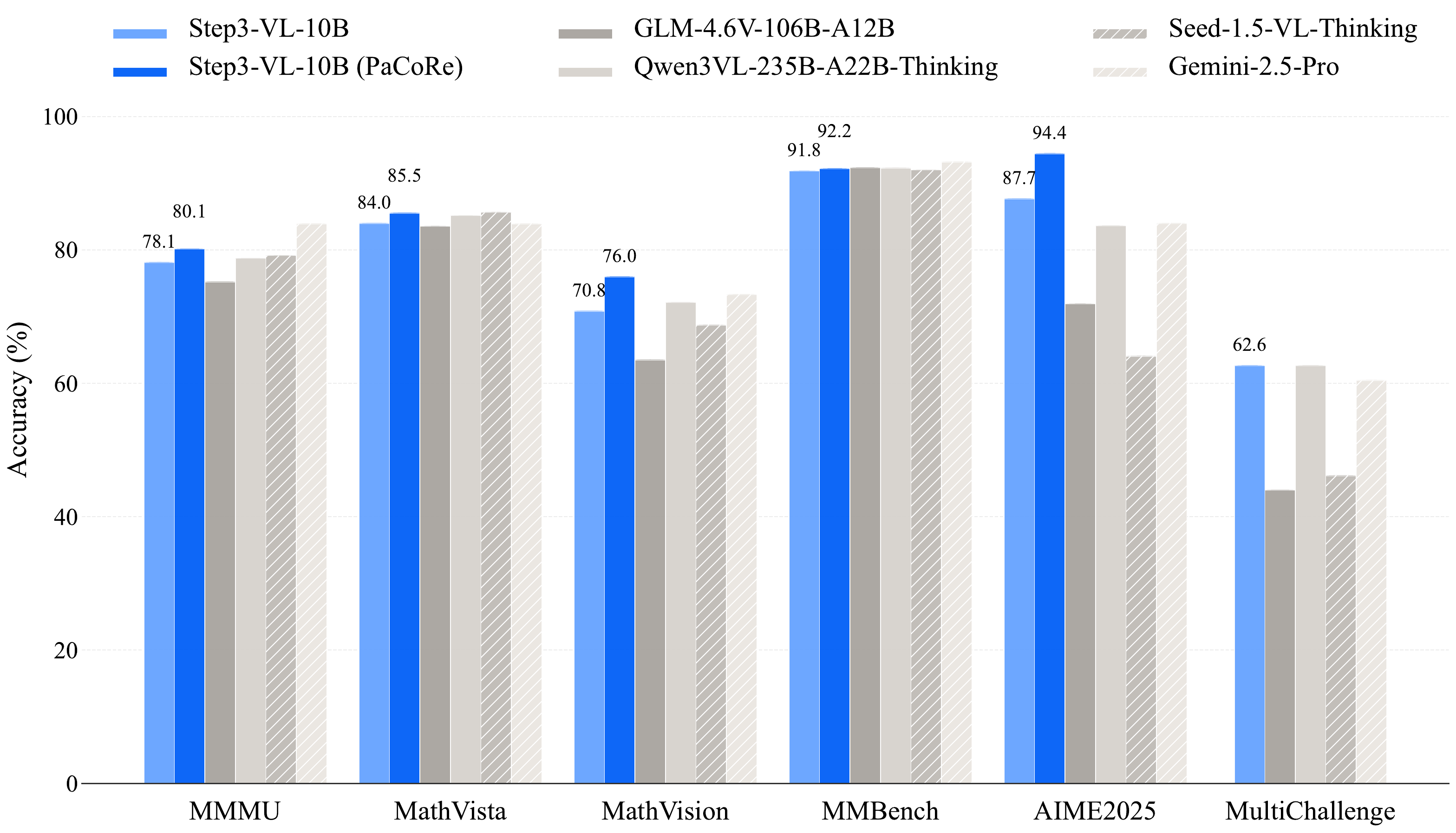}
\caption{\textbf{Performance comparison of \modelname~against state-of-the-art multimodal foundation models.} With PaCoRe (Parallel Coordinated Reasoning~\citep{hu2025pacore}, \modelname~scales \textbf{test-time compute} to bridge the perception and reasoning performance gap with \textbf{100B+ parameter models}.}
\label{fig:preview}
\end{figure}
\vspace{0.5em}

\definecolor{colorfirst}{RGB}{252,141,89}
\definecolor{colorsecond}{RGB}{253,187,132}
\definecolor{colorthird}{RGB}{253,212,158}
\definecolor{colorfourth}{RGB}{254,232,200}
\definecolor{colorfifth}{RGB}{255,247,236}
\definecolor{myred}{RGB}{242,128,128}
\definecolor{mygreen}{RGB}{112,180,143}
\definecolor{myblue}{RGB}{210,225,255}
\definecolor{citypink}{RGB}{227,108,194}
\definecolor{cityblue}{RGB}{128,159,225}
\definecolor{casepromptlightblue}{HTML}{66a3ff}
\definecolor{caseanswerdarkblue}{HTML}{025ff7}
\newcommand{\rankfirst}[0]{\cellcolor{colorfirst}}
\newcommand{\ranksecond}[0]{\cellcolor{colorsecond}}
\newcommand{\rankthird}[0]{\cellcolor{colorthird}}
\newcommand{\rankfourth}[0]{\cellcolor{colorfourth}}
\newcommand{\rankfifth}[0]{\cellcolor{colorfifth}}
\DeclareRobustCommand{\legendsquare}[1]{%
  \textcolor{#1}{\rule{2ex}{2ex}}%
}
\DeclareRobustCommand{\legendsquarebox}[1]{%
  \tikz[] \draw[black, fill=#1, line width=0.4pt] (0,0) rectangle (1.5ex,1.5ex);%
}
\newcommand{\cmark}{\textcolor{mygreen}{\ding{51}}}%
\newcommand{\xmark}{\textcolor{myred}{\ding{55}}}%

\definecolor{casebglight}{RGB}{245, 247, 250}
\definecolor{caseborderblue}{RGB}{64, 158, 255}

\newtcolorbox{caseprompt}{
    colback=casebglight,
    colframe=casepromptlightblue,
    title=\textbf{User Prompt},
    fonttitle=\bfseries,
    boxrule=0.5mm,
    sharp corners,
    left=2mm, right=2mm, top=2mm, bottom=2mm
}

\newtcolorbox{casethinking}{
    colback=white,
    colframe=gray!50,
    title=\textbf{Model Thinking Process},
    fonttitle=\bfseries,
    boxrule=0.3mm,
    sharp corners,
    left=2mm, right=2mm, top=2mm, bottom=2mm,
    coltitle=black,
    attach title to upper={\par\vspace{2mm}},
    breakable
}

\newtcolorbox{caseanswer}{
    colback=casebglight,
    colframe=caseanswerdarkblue,
    title=\textbf{Final Answer},
    fonttitle=\bfseries,
    boxrule=0.5mm,
    sharp corners,
    left=2mm, right=2mm, top=2mm, bottom=2mm
}

\newpage

\begin{spacing}{0.9}
\tableofcontents
\end{spacing}

\newpage

\section{Introduction}

The development of Multimodal Large Language Models (MLLMs) has largely been driven by a relentless pursuit of scale. While proprietary frontier models like Gemini-3-Pro~\citep{gemini3pro} and GPT-5.2~\citep{gpt5p2} have pushed the boundaries of \textit{\textbf{multimodal intelligence}} through massive scaling, their heavy computational demands pose barriers to practical deployment in the real world. Conversely, lightweight models (under 10B parameters) have traditionally been characterized as ``efficient but limited'', which struggle to advance sophisticated reasoning and perceptual capabilities within restricted parameter budgets.

In this work, we introduce \textbf{\modelname}, a foundation model that redefine the trade-off between compact efficiency and frontier-level multimodal intelligence. Despite its modest 10B parameter footprint, \modelname~excels in \textbf{visual perception}, \textbf{complex reasoning}, and \textbf{human-centric alignment}. It consistently outperforms models under the 10B scale  and rivals or even surpasses significantly larger open-weights models (\textbf{10$\times$--20$\times$ its size}), such as GLM-4.6V (106B-A12B)\citep{glm4p6v} and Qwen3-VL-Thinking (235B-A22B)\citep{bai2025qwen3vltechnicalreport}, as well as \textbf{established proprietary flagships} like Gemini-2.5-Pro\citep{gemini2p5pro} and Seed-1.5-VL~\citep{seed1p5vl}. Across representative benchmarks, \modelname~achieves 75.95\% on MathVision, 80.11\% on MMMU, and a staggering 94.43\% on AIME2025 (\Cref{fig:preview}).

The success of \modelname~is driven by two key strategic design in how we build efficient and powerful multimodal models:


\begin{itemize}
    \item \textbf{Unified Pre-training on High-Quality Multimodal Corpus:} 
    We implement a \textbf{single-stage, fully unfrozen training strategy} on a 1.2T token multimodal corpus, focusing on two foundational capabilities: \textbf{reasoning} (e.g., general knowledge and education-centric tasks) and \textbf{perception} (e.g., grounding, counting, Optical Character Recognition, and Graphical User Interface interactions). By jointly optimizing the 
    Perception Encoder~\citep{bolya2025perception} and the Qwen3-8B~\citep{yang2025qwen3} decoder, \modelname~establishes a \textbf{intrinsic vision-language synergy}.
    \item \textbf{Scaled Multimodal Reinforcement Learning (RL) and Parallel Reasoning:} We unlock frontier capabilities through a rigorous post-training pipeline, comprising two-stage supervised finetuning (SFT) and \textbf{over 1k iterations of RL} with both verifiable rewards (RLVR) and human feedback (RLHF). Beyond sequential reasoning, we adopt Parallel Coordinated Reasoning (PaCoRe)~\citep{hu2025pacore}, which allocates test-time compute to \textbf{aggregate evidence from parallel visual exploration}. These designs enable the 10B model to solve complex perceptual and reasoning tasks that typically require substantially larger systems.

\end{itemize}

To understand the drivers of this efficiency, we provide a in-depth analysis of the model's internal mechanisms in Sec.\ref{sec:discuss}, with an emphasis on the learning dynamics unlocked by RL scaling. 
In particular, to counteract the length diminishment characteristic of perception tasks, we leverage PaCoRe to facilitate a form of multi-agent synthesis: parallel proposers generate diverse hypotheses, which are subsequently distilled through sequential cross-checking. This emergent synthesis effectively externalizes implicit visual processes, offering a promising direction for \textbf{scaling perceptual reasoning}.


This trajectory not only sheds light on how models can progressively bridge intelligence and interaction with the physical world (Sec.\ref{sec:limitations}), but also motivates our commitment to closing critical technical gaps in the open ecosystem. By releasing the final model weights and detailed training documentation, \modelname~demonstrates that with a right perception- and reasoning-centric design, the gap between ``compact'' and ``frontier'' is no longer intractable.


\setlength{\epigraphwidth}{0.95\columnwidth}
\renewcommand{\epigraphflush}{center}
\renewcommand{\textflush}{flushepinormal}
\renewcommand{\epigraphsize}{\footnotesize}





\section{Pre-train}
\label{sec:pretrain}
Our pre-training framework is designed to construct a capable vision foundation model that sets a high upper bound for subsequent post-training stages, prioritizing data quality and architectural synergy over unnecessary complexity.

\subsection{Architecture}

\modelname~integrates the 1.8B language-optimized Perception Encoder~\citep{bolya2025perception}, selected over the spatial-optimized variant for its pre-aligned linguistic features that ensure superior convergence. This visual backbone is coupled with Qwen3-8B~\citep{yang2025qwen3}, utilized as the decoder for its robust text generation foundation and proven plasticity for multimodal adaptation. 
Similar to Step-3~\citep{stepfun2025step3} and DeepSeek-OCR~\citep{wei2025deepseekocr}, these components are bridged by a projector performing $16\times$ spatial downsampling via two consecutive stride-2 layers, effectively compressing visual tokens while retaining essential information. To capture fine-grained details efficiently, we adopt a multi-crop strategy~\citep{caron2021multicrop} that decomposes images into a $728\times728$ global view and multiple $504\times504$ local crops. This design leverages batch-dimension parallelism to sidestep the complexity of variable-length packing~\citep{shah2024flashattention}. Finally, we encode spatial structure by appending newline tokens to patch rows and utilize standard 1D RoPE~\citep{su2024roformer} for positional modeling, as advanced variants yielded no significant gains in our setup.

\subsection{Data Construction}
To equip \modelname~with both fine-grained perception and complex reasoning capabilities, we incorporate a large-scale text corpus~\citep{bakouch2025smollm3} and construct a comprehensive multimodal pre-training dataset spanning the following key domains.


\paragraph{Knowledge.}
We curate visual data with high knowledge density from multiple complementary channels, covering both structured interleaved data and diverse image–text pairs.

\begin{itemize}
    \item \textbf{Interleaved Data.}
    We collect interleaved image–text data from \textbf{Common Crawl}~\citep{commoncrawl} and our in-house crawler \textbf{StepCrawl}, which targets the domestic internet, and further augment this corpus with keyword-based search results. To suppress noise inherent in web-scale data, we discard webpages with excessive image download failures ($>90\%$), QR-code content, and images with extreme aspect ratios.

    \item \textbf{Image--Text Pairs.}
    We organize image–text pairs into four complementary categories:
    (1) \textbf{Open-source datasets}, including LAION~\citep{laion5b}, COYO~\citep{coyo}, BLIP-CCS~\citep{blip}, and Zero~\citep{Xie2023zero}. To mitigate long-tail concept imbalance, we apply \textit{concept-balanced resampling} via CLIP~\citep{radford2021clip}-based clustering.
    (2) \textbf{Keyword-based retrieval}, where keywords mined from high-quality knowledge websites are used to query commercial search engines (e.g., Baidu and Bing), to gather targeted domain-specific data.
    (3) \textbf{Pairs extracted from interleaved data}, where for each image we extract candidate descriptions from surrounding text (above, below, and alt-text). Then we select the most suitable one using CLIP-based similarity to assess alignment and aesthetic scores to evaluate image quality.
    (4) \textbf{Mosaic augmentation}, in which four images are concatenated into a single input. This effectively extends the input resolution, increases visual content density within each sample, and encourages the model to learn spatial and positional reasoning across multiple regions.
\end{itemize}

\paragraph{Education.} We curate a dataset of approximately 15M samples spanning K-12 education, higher education, and adult learning. 
The \textbf{K-12} subset covers mathematics, physics, chemistry, and humanities, and includes specialized data such as chemical formulas and structure diagrams sourced from open datasets and synthetically generated using CoSyn~\citep{yang2025cosyn}, as well as geometry (including analytic geometry) problems constructed from a mixture of synthetic data and real exam images with annotated captions. 
Beyond this, the dataset extends to \textbf{university-level} domains including STEM, medicine, arts, and finance, as well as \textbf{adult education} scenarios such as driving license exams, CPA, and legal examinations. 
Exam questions are collected from a combination of licensed exam materials and open-source problem sets~\citep{sujetfinanceqavision100k, vqamed, pathvqa}, while supporting knowledge content is sourced from textbooks, workbooks, and high-quality educational websites.

\paragraph{Optical Character Recognition (OCR).} We curate a comprehensive OCR corpus spanning image-level and document-level text recognition, and visual-to-code reconstruction.
\begin{itemize}
    \item \textbf{Image to Text.} We curate a dataset comprising 10M real-world images and 30M synthetic samples covering diverse fonts, layouts, and text orientations. Real-world data are collected from open-source datasets~\citep{shi2017icdar2017, cong2012rects} and annotated using PaddleOCR~\citep{cui2025paddleocr3}, while synthetic samples are generated using SynthDog~\citep{kim2022synthdog}.

    \item \textbf{Image to Code.} We organize this dataset by target code form, spanning markup-based and programmatic graphics.
    (1) \textbf{Markup-based Code.}
    For Markdown, \LaTeX{}, and Matplotlib, we combine over 10M samples from open-source datasets~\citep{unichartqa,simchart9k,chen2024onechart} with an automated data generation pipeline that produces more than 15M synthetic infographics. Instead of fully delegating generation to LLMs~\citep{yang2025cosyn}, we enforce fine-grained rendering rules across multiple render tools.
    (2) \textbf{Programmatic Graphics Code.}
    For languages such as TikZ and Graphviz, we curate approximately 5M reconstruction tasks that require translating visual inputs into executable code, spanning diverse visual inputs including natural images, human-created tables, and geometries.
    
    \item \textbf{Document to Text.} This dataset comprises approximately 80M full-page documents. Concretely, we apply PaddleOCR or MinerU 2.0~\citep{wang24mineru} to annotate collected pages from books and academic papers.
    
    \item \textbf{Document to Code.} We curate data spanning three primary markup languages including HTML, Markdown and latex. \textbf{HTML} data focus on table-centric content and are sourced from rendered web page code, Markdown conversions. \textbf{Markdown} data cover tables and lightweight documents, collected from rendered GitHub README files, HTML conversions. \textbf{Latex} data are extracted at scale from arXiv corpora, comprising approximately 4M tables and 100M formulas. During rendering, we explicitly handle elements such as references and hyperlinks to prevent mismatches between visual and textual content. In addition, we incorporate open-source datasets~\citep{doclatex,hme100k,websight,liu2024focus} to further enrich this subset.

\end{itemize}

\paragraph{Grounding \& Counting.} We collect approximately 400M samples to support fine-grained perceptual understanding. \textbf{Grounding data} include both bounding-box-based and point-based annotations sourced from open detection datasets such as OpenImages~\citep{Kuznetsova2020openimage}, COCO~\citep{coco}, Merlin~\citep{yu2024merlin,yu2025unhackable}, and PixMo~\citep{deitke2024pixmo}, as well as in-house text paragraph detection tasks. \textbf{Counting data} are drawn from open-source counting~\citep{fcs,locount} and are further constructed by converting high-quality object detection annotations into counting formulations.

\paragraph{Visual Question Answering (VQA).}
This subset comprises approximately 10M samples targeting holistic image content understanding. It includes curated open-source VQA datasets~\citep{liu2025conflict,zellers2019vcr}, as well as high-quality question–answer pairs automatically generated from image caption data.
In addition, we construct a OCR VQA subset with around 20M samples, combining open-source data~\citep{olmocrmix,nyubookevaleg2,wei2024general} with QA pairs generated from other OCR-related task.

\paragraph{Graphical User Interface (GUI).}
We construct a large-scale GUI dataset as in \textit{Step-GUI}~\citep{yan2025step} comprising approximately 23M samples to endow the model with practical and executable UI understanding and interaction capabilities. The dataset covers both mobile platforms, including Android and iOS, and desktop environments spanning Windows, Linux, and macOS, with data collected from over 200 applications. Notably, accurate grounding annotations are generated jointly with trajectory data, ensuring consistent supervision between action execution and perception.

\begin{itemize}
    \item \textbf{Caption.} This subset provides 700K detailed captions for UI interfaces, conveying explicit knowledge about page layouts and functional regions to support structural understanding of interfaces.

    \item \textbf{Knowledge VQA.} We include over 1M question–answer pairs that reinforce precise localization and functional understanding of UI elements.
    
    \item \textbf{Trajectory Modeling.} To model realistic sequential human–computer interactions, we curate more than 2M trajectory samples defined over a granular action space comprising 12 atomic actions, such as \texttt{CLICK}, \texttt{SLIDE}, and \texttt{TYPE}. These trajectories strengthen action output formatting, state summarization, and decision-making capabilities, and cover a wide range of tasks including operation execution, information retrieval, and information summarization.

    \item \textbf{Grounding.} The dataset further includes over 19M grounding samples with both point-based and bounding-box-based grounding across diverse interface layouts and resolutions.
    
    \item \textbf{OCR.} For web-based interfaces targeting Browser-use-GUI, we crawl approximately 30M web pages and extract textual content together with precise element coordinates, supporting fine-grained layout-aware understanding.

\end{itemize}

\subsection{Training Recipe}
We adopt a single-stage, fully unfrozen training strategy optimized with AdamW~\citep{loshchilov2017adamw} ($\beta_1 = 0.9$, $\beta_2 = 0.95$, $\epsilon = 10^{-8}$, and weight decay $= 0.01$), training the model on a total of 1.2T tokens over 370K iterations with a global batch size of 8{,}192 and a sequence length of 4{,}096.

To balance training scale and data quality, we employ a two-phase learning rate schedule. During the first phase, covering the initial 900B tokens, the learning rate is decayed from $5 \times 10^{-5}$ to $1 \times 10^{-5}$ to emphasize broad representation learning. In the second phase, spanning the remaining 300B tokens, we transition to a higher-quality data mixture and further anneal the learning rate from $1 \times 10^{-5}$ to $6 \times 10^{-6}$,
acting as a cool-down phase to consolidate fine-grained perceptual (e.g., OCR and grounding) and reasoning capabilities .

\section{Post-Train}

In the post-training stage, we adopt a two-stage Supervised Finetuning (SFT) strategy followed by Reinforcement Learning (RL). For the RL phase, we employ Proximal Policy Optimization (PPO)~\citep{ppo_cite} with Generalized Advantage Estimation (GAE)~\citep{gae_cite} as the core optimization algorithm, coupled with a meticulously designed reward system. Crucially, we scale inference compute by progressing from sequential reasoning to parallel coordinated reasoning, aiming to fully unlock \modelname's perception and reasoning capabilities.

\label{sec:posttrain}
\subsection{Supervised Finetuning}

\paragraph{Data Construction.}
Our SFT strategy focuses on multi-modal, high-quality, reasoning-oriented data. We initially collected millions of prompts from the open-source community~\citep{guha2025openthoughtsdatarecipesreasoning,numina_math_datasets}, spanning diverse domains such as mathematics, coding, science, and logical reasoning. We also incorporated open-source datasets~\citep{wiedmann2025finevision,tong2024cambrian,onevision} for visual perception and recognition, including grounding, OCR, and complex document/chart understanding, to ensure the model can precisely perceive and reason over visual elements. Leveraging these prompts, we distilled high-quality responses from internal frontier model. This foundational dataset underwent a rigorous ``two-pipe'' filtration process: first, applying predefined rules to eliminate degenerate patterns (e.g., infinite repetitions); and second, performing comprehensive benchmark decontamination via exact matching and $N$-gram matching ($N=64$).

\paragraph{Two-Stage SFT Strategy.}
We implemented a phased training approach to progressively align the model's reasoning capabilities across modalities. The training was conducted with a global batch size of $32$ and an extended sequence length of $128$k to support long-context understanding.
\begin{itemize}
    \item \textbf{Stage 1: Text-Dominant Reasoning.} The data mixture was set at a $9:1$ ratio of text to multimodal samples, establishing a strong logical and linguistic foundation.
    \item \textbf{Stage 2: Multimodal Integration.} We shifted the composition to a $1:1$ ratio, effectively balancing textual reasoning with visual intelligence to enhance the model's performance on interleaved multimodal tasks.
\end{itemize}

\paragraph{Training Recipe.}
We employed a cosine learning rate scheduler with a $200$-step warmup phase, where the learning rate peaks at $1 \times 10^{-4}$ and anneals to a final value of $1 \times 10^{-5}$. To optimize the learning process across diverse data sources, we implemented domain-specific sampling weights in the dataloader, which translate to varying epoch counts for different domains. Throughout the entire two-stage process, the model was trained on a total of approximately $190$B tokens during stage 1 and $36$B during stage 2.

\subsection{Reinforcement Learning}
\subsubsection{Optimization Algorithm}

We adopt PPO combined with GAE as our optimization algorithm for reinforcement learning, following \textit{Open-Reasoner-Zero}~\citep{hu2025open} and \textit{Open-Vision-Reasoner}~\citep{ovr}.

Formally, given a multimodal input tuple consisting of an image $I$ and a textual prompt $q$, the policy network $\pi_\theta$ generates a response trajectory $\tau = (s_0, a_0, \ldots, s_{T-1}, a_{T-1})$ of length $T$. The state $s_t$ encapsulates the input context $(I, q)$ and the sequence of tokens generated prior to step $t$, while $a_t$ denotes the action (token) sampled at step $t$.

To effectively balance the bias-variance trade-off in policy gradient estimation, we utilize GAE for advantage computation. The advantage estimator $\hat{A}_t$ for a state-action pair $(s_t, a_t)$ is defined as:
\begin{equation}
\hat{A}_t = \sum_{l=0}^{T-t-1} (\gamma \lambda)^l \delta_{t+l}, \quad \text{with} \quad \delta_{t'} = r_{t'} + \gamma V_\phi(s_{t'+1}) - V_\phi(s_{t'}),
\end{equation}
where $r_{t'}$ is the reward at step $t'$, $V_\phi$ represents the value function parameterized by $\phi$, and $\gamma, \lambda \in [0, 1]$ are the discount factor and GAE smoothing parameter, respectively.

The policy parameters $\theta$ are updated by maximizing the clipped surrogate objective, which penalizes large policy deviations to maintain training stability:
\begin{equation}
\mathcal{J}_{\text{PPO}}(\theta) = \hat{\mathbb{E}}_{t} \left[ \min \left( \rho_t(\theta) \hat{A}_t, \text{clip} \left( \rho_t(\theta), 1 - \epsilon, 1 + \epsilon \right) \hat{A}_t \right) \right],
\end{equation}
where $\rho_t(\theta) = \frac{\pi_\theta(a_t | s_t)}{\pi_{\text{old}}(a_t | s_t)}$ is the probability ratio, and $\epsilon$ is the clipping hyperparameter. 


Concurrently, the value function is updated to minimize the mean squared error between the estimated value and a target value $V_t^{\text{target}}$, typically the estimated discounted return $G_t = \hat{A}_t^{\text{GAE}(\gamma, \lambda)} + V_\phi(s_t)$:
\begin{equation}
\mathcal{J}_{\text{value}}(\phi) = \frac{1}{2} \mathbb{E}_{\tau \sim \pi_{\theta_{\text{old}}}} \left[ \sum_{t=0}^{T-1} \left( V_\phi(s_t) - V_t^{\text{target}} \right)^2 \right],
\end{equation}

Concretely, we adopt a variant of PPO algorithm with GAE ($\gamma = 1, \lambda = 1$) for off-policy setting, omitting standard importance sampling. Each iteration splits samples into four mini-batches. The actor and critic learning rates are set to $2 \times 10^{-6}$ and $5 \times 10^{-6}$, respectively. To mitigate training--inference inconsistency, we apply the truncated importance sampling ratio with threshold $C = 8$ following~\citep{yao2025offpolicy}. The entire RL phase runs for 1,400 training iterations, updating only the decoder while keeping the encoder frozen.

\subsubsection{Reward System}

To support scalable training across heterogeneous modalities and task types, we design a bifurcated reward framework that explicitly distinguishes between \textit{verifiable tasks}, where objective correctness can be reliably assessed, and \textit{non-verifiable tasks}, where alignment must be guided by preference modeling and constraints. 

\paragraph{Verifiable Rewards: Precision and Consistency.}
For tasks with accessible ground truth, our reward design prioritizes strict correctness and reasoning consistency. We implement a multi-faceted verification pipeline that combines perception-based, and model-assisted signals.

\begin{itemize}
    \item \textbf{Perception Rewards.} 
    For perception tasks such as pointing and grounding, following \textit{Perception-R1}~\citep{yu2025perception}, we align the model's geometric outputs with deterministic ground truths using metrics like Intersection over Union (IoU) or Euclidean distance. Crucially, we implement strict, distance-decay reward shaping to guarantee a distinct and unambiguous optimization landscape and robust RL convergence.
    \item \textbf{Model-Based Verification.} For general visual tasks, we deploy \textbf{GPT-OSS-120B}~\citep{openai2025gptoss} as the answer verifier. Compared to simple string matching or \textit{mathverify}-style heuristics, this model-based verification mechanism is substantially more robust to noisy or imperfect ground truth and significantly improves training stability. In particular, it provides \textbf{parse-invariant} evaluation that is resilient to formatting variations (e.g., idiosyncratic \LaTeX), recognizes \textbf{semantic equivalence} among mathematically identical expressions or reordered derivation steps, and enforces \textbf{process consistency} by penalizing false positives, cases where the model arrives at a correct final answer through flawed or unsupported reasoning. Together, these properties yield more reliable reward signals for supervising complex reasoning behaviors.
\end{itemize}

\paragraph{Non-Verifiable Rewards: Preference and Constraints.}
For open-ended generation where ground truth is absent, we rely on learned preference models and heuristic constraints to guide human-centric alignment.
\begin{itemize}
    \item \textbf{Generative Reward Modeling (GenRM).} We adopt a pairwise preference framework where the GenRM evaluates rollouts against responses generated by a more capable teacher model. Moving beyond direct outcome continuous rewarding, our GenRM incorporates an explicit reasoning judgment before deriving a fine-grained scalar score to discern subtle quality differences between plausible responses.
    \item \textbf{Behavioral Regularization.} To mitigate ``reward hacking'' and enforce safety and reliability constraints during optimization, we incorporate a set of model-based penalty terms as behavioral regularization. Specifically, we impose \textbf{language consistency} penalties to discourage code-switching and question–answer language mismatch; apply strict \textbf{citation verification} that zeros the reward when fabricated references or links are detected, directly targeting hallucinations at the source; and introduce \textbf{epistemic calibration} penalties to suppress unjustified certainty or overconfident claims, encouraging the model to appropriately express uncertainty in ambiguous or underspecified settings. Together, these constraints act as guardrails that stabilize preference optimization and align model behavior with safety and trustworthiness objectives.
\end{itemize}

\subsubsection{Scaling Sequential Reasoning}

We aim to scale the model's reasoning capability by incentivizing extended sequential thinking processes, effectively converting test-time compute into performance gains. Concretely, we structure our sequential reasoning training to first establish robust logical foundations on verifiable tasks before aligning with subjective human preferences.

\paragraph{Reinforcement Learning with Verifiable Rewards (RLVR).} 
We conduct training on a diverse set of verifiable multimodal tasks, drawing from large-scale open-source datasets such as \textit{Open-Vision-Reasoner}~\citep{ovr}, which cover mathematics, geometry, physics, scientific reasoning, perception, recognition, chart-based reasoning, and puzzles, together with visual grounding tasks from \textit{Perception-R1}~\citep{yu2025perception} and internal K–12 educational resources. 

To ensure high-quality supervision, we design a multi-dimensional filtration pipeline along three axes: 
\textbf{(1) Checkability} is enforced by employing GPT-OSS-120B to perform four independent verification passes per prompt, retaining only \textit{all-agree} samples.
\textbf{(2) Visual relevance} is ensured by using an early version of \modelname~to evaluate the semantic correlation and mutual contribution between images and questions, filtering out redundant or misaligned multimodal pairs. 
\textbf{(3)} Finally, to control \textbf{difficulty}, we perform 24 rollouts per prompt to identify \textit{some-accept} samples, namely cases that are neither trivially solvable nor consistently failed.
Each filtration stage plays a non-trivial role in ensuring long-term training stability and enabling sustained performance gains.
The RLVR stage is executed for 600 iterations with a maximum sequence length of 24k. For each iteration, we sample 512 prompts with 16 rollouts per prompt, optimizing via the aforementioned verifiable reward system.

\paragraph{Reinforcement Learning from Human Feedback (RLHF).} Building on the reasoning-focused checkpoint from RLVR, we further align the model with human preferences using open-ended tasks. We curate prompts from opensourced arena datasets~\citep{chou2024visionarena,tang2025explorer,chiang2024chatbot} and internal instruction pools, explicitly filtering for \textit{uncheckable} queries that lack deterministic ground-truth. For these prompts, we leverage our strongest internal models
to generate high-quality reference responses as anchors for preference learning. This stage proceeds for 300 iterations using a maximum sequence length of 32k. We maintain a throughput of 512 prompts per iteration with 8 rollouts per prompt, optimizing the model against an unverifiable reward system to refine its conversational and alignment capabilities while preserving its underlying reasoning strength.

\subsubsection{Further Scaling Parallel Coordinated Reasoning}

To further scale test-time compute beyond the limits of sequential generation, we adopt a \textbf{parallel coordinated reasoning} paradigm following \textit{PaCoRe}~\citep{hu2025pacore}. This approach allocates compute to explore diverse perceptual hypotheses in parallel and synthesizes them into a unified conclusion.

To construct the training data for parallel reasoning, we extend the difficulty filtration (Axis 3) from the RLVR stage. We repurpose the 24 rollouts from the difficulty tagging phase as a \textit{message cache pool}. Starting with the identified \textit{some-accept} prompts, we apply a secondary Synthesis Filtration to ensure \textbf{coordinated solvability}: (1) We simulate the parallel reasoning process by sampling 16--24 messages from the pool and feeding them back into the model as a "synthesis context" to regenerate responses. (2) We strictly retain instances that \textit{remain} categorized as \textit{some-accept} 
under this coordinated setting.
Crucially, this prevents task trivialization to maintain effective reward signals, while compelling the model to perform multi-perspective self-verification and cross-checking.


The model is optimized using PPO in a strict on-policy setting for 500 iterations. We utilize a maximum sequence length of 64k to accommodate the aggregated context. Each iteration samples 64 prompts with 16 rollouts per instance, stabilizing the optimization of coordinated behaviors against the verifiable reward system.




\section{Evaluations}
\label{sec:evaluations}

To rigorously validate the capabilities of \modelname, we conduct extensive evaluations across a broad spectrum of multimodal and text-centric benchmarks. Our results position \modelname~as the \textbf{most powerful open-source model} in the 10B parameter class, demonstrating performance that not only significantly outperforms \textbf{7--10B open-source models} but also rivals frontier open-source systems (10$\times$--20$\times$ larger) and proprietary flagships in reasoning and perception domains.


\subsection{Evaluation Setup}

\paragraph{Benchmark Protocols.} 
We evaluate \modelname~on a comprehensive suite of over 60 benchmarks. To ensure a holistic assessment consistent with our reported results, we categorize these benchmarks into specific capability domains across multimodal and text-centric modalities.

\textbf{I. Multimodal Benchmarks.} We assess vision-language capabilities across nine distinct domains:
\begin{itemize}
    \item \textbf{STEM / Multimodal Reasoning:} We employ MMMU (Standard/Pro)~\citep{yue2024mmmu}, MathVista~\citep{lu2023mathvista}, MathVision~\citep{wang2024measuringmultimodalmathematicalreasoning}, MathVerse~\citep{zhang2024mathversedoesmultimodalllm}, DynaMath~\citep{zou2024dynamic}, We-Math~\citep{qiao2024we}, and PhyX~\citep{phyx} for scientific and mathematical reasoning. Logical and puzzle-solving abilities are tested via LogicVista~\citep{xiao2024logicvistamultimodalllmlogical}, ZeroBench~\citep{roberts2025zerobenchimpossiblevisualbenchmark}, VisuLogic~\citep{xu2025visulogicbenchmarkevaluatingvisual}, and HLE~\citep{phan2025humanitysexam} .
    
    \item \textbf{Recognition / General VQA:} General perception is evaluated using MMBench (EN/CN)~\citep{mmbench}, SimpleVQA~\citep{cheng2025simplevqamultimodalfactualityevaluation}, and MMStar~\citep{mmstar}. Robustness and fine-grained recognition are assessed via HallusionBench~\citep{hallusionbench}, MMVP~\citep{mmvp}, ReMI~\citep{kazemi2024remidatasetreasoningmultiple}, M3GIA~\citep{song2024m3giacognitioninspiredmultilingual}, and DoYouSeeMe~\citep{kanade2025multidimensionalbenchmarkevaluating}.
    
    \item \textbf{Counting:} We utilize CountBench~\citep{paiss2023teachingclipcount}, CountQA~\citep{tamarapalli2025countqamllmscountwild}, and PixMo-Count~\citep{deitke2024pixmo} to evaluate precise object enumeration.
    
    \item \textbf{Instruction Following:} Multimodal compliance is tested on MM-MT-Bench~\citep{ying2024mmtbenchcomprehensivemultimodalbenchmark}, MIA-Bench~\citep{qian2025miabenchbetterinstructionfollowing}, and MM-IFEval~\citep{mmifeval}.
    
    \item \textbf{Multimodal Code:} Visual coding capabilities are evaluated using HumanEval-V~\citep{humanevalv}, and Design2Code (including Design2Code-Hard)~\citep{design2code}.
    
    \item \textbf{OCR:} Text-rich image understanding is assessed via OCRBench~\citep{ocrbench}, OmniOCR~\citep{omniocr}, and CC-OCR~\citep{ccocr}.
    
    \item \textbf{2D / 3D Spatial Understanding:} We conduct extensive spatial reasoning tests using BLINK~\citep{blink}, CVBench~\citep{tong2024cambrian1}, MMSI-Bench~\citep{yang2025mmsibenchbenchmarkmultiimagespatial}, ERQA~\citep{geminiroboticsteam2025geminiroboticsbringingai}, OmniSpatial~\citep{jia2025omnispatialcomprehensivespatialreasoning}, All-Angles-Bench~\citep{yeh2025seeingperspectiveevaluatingmultiview}, MindCube-tiny~\citep{yin2025spatialmentalmodelinglimited}, RealWorldQA~\citep{realworldqa}, SpatialViz-Bench~\citep{wang2025spatialviz}, STARE~\citep{stare}, CoreCognition~\citep{li2025coreknowledgedeficitsmultimodal}, V*~\citep{vstar}, and ViewSpatial~\citep{li2025viewspatialbenchevaluatingmultiperspectivespatial}.
    
    \item \textbf{Document \& Chart Understanding:} Complex parsing is tested on CharXiv (RQ)~\citep{charxiv}, AI2D~\citep{ai2d}, OmniDocBench~\citep{omnidocbench}, and 
    CSVQA~\citep{csvqa}.
    
    \item \textbf{GUI Grounding:} To evaluate actionable intelligence, we employ ScreenSpot-Pro~\citep{screenspot}, ScreenSpot-V2~\citep{wu2024atlas}, OSWorld-G~\citep{osworldg}, and MMBench-GUI~\citep{mmbenchgui}.
\end{itemize}

\textbf{II. Text-Centric Benchmarks.} We verify LLM foundations across six categories:
\begin{itemize}
    \item \textbf{Exam:} General knowledge is evaluated on MMLU-Pro~\citep{mmlu_pro_cite}, GPQA-Diamond~\citep{rein2023gpqa}, SuperGPQA~\citep{supergpqa}, and LiveBench~\citep{livebench}.
    
    \item \textbf{Mathematics:} Mathematical reasoning is rigorously tested on AIME (2024/2025)~\citep{aime2024, aime2025}, Beyond-AIME~\citep{beyondaime}, HMMT25~\citep{hmmt2025}, CNMO 2024~\citep{cnmo}, and IMO-AnswerBench~\citep{luong2025robustmathematicalreasoning}.
    
    \item \textbf{Code:} Pure text coding is evaluated via LiveCodeBench (2408-2505)~\citep{jain2024livecodebench}.
    
    \item \textbf{Instruction Following:} We use IFEval~\citep{ifeval}, IFBench~\citep{ifbench}, and MultiChallenge~\citep{multichallenge}.
    
    \item \textbf{Subjective:} Open-ended generation quality is assessed on Arena-Hard-V2~\citep{arenahard} and WildBench~\citep{wildbench}.
    
    \item \textbf{Medical:} Domain-specific knowledge is tested on HealthBench~\citep{healthbench}.
\end{itemize}

\paragraph{Inference Settings.}
We evaluate \modelname~using a fixed configuration (temperature=1.0, top-p=1.0, top-k=0). By default, the model uses \textbf{Sequential Reasoning (SeRe)}, generating thoughts wrapped in \texttt{<think>} and \texttt{</think>} tags before the answer, with a maximum length of 65{,}536 tokens. For complex perception and advanced reasoning tasks, we employ \textbf{Parallel Coordinated Reasoning (PaCoRe)}~\citep{hu2025pacore}, which synthesizes 16 SeRe rollouts into a context for final inference (more details refer to \Cref{sec:synthesis}). In PaCoRe mode, the maximum length is extended to 131{,}072 tokens to support the expanded context, while other hyperparameters remain consistent.


\paragraph{Comparison Models.}
We benchmark \modelname~against representative open-source models (7B--10B) including \textbf{GLM-4.6V-Flash} (9B)~\citep{glm4p6v}, \textbf{Qwen3-VL-Thinking} (8B)~\citep{bai2025qwen3vltechnicalreport}, \textbf{InternVL-3.5} (8B)~\citep{wang2025internvl3p5}, and \textbf{MiMo-VL-RL-2508} (7B)~\citep{mimovl}. For scalability analysis, we compare against larger systems: \textbf{GLM-4.6V} (106B-A12B)~\citep{glm4p6v}, \textbf{Qwen3-VL} (235B-A22B)~\citep{bai2025qwen3vltechnicalreport}, \textbf{Gemini-2.5-Pro}~\citep{gemini2p5pro}, and \textbf{Seed-1.5-VL}~\citep{seed1p5vl}.


\subsection{Multimodal Evaluation Results}

In Table~\ref{tab:same_size_bmk_mm}, we benchmark \modelname~against strong open-source models within the 7B--10B parameter range. The results indicate that \modelname~establishes a new performance standard for compact models, securing the top position in almost all capability domains. We provide a detailed breakdown below.

\begin{table}[htbp]
  \centering
  \scriptsize
  \caption{Comparison with state-of-the-art open-source models (7B--10B) on multimodal benchmarks. The \textbf{bold} and \underline{underlined} numbers indicate the best and second-best results, respectively. $^*$ indicates results reported in the original papers.}
  \label{tab:same_size_bmk_mm}
  \setlength{\tabcolsep}{4pt}
  \renewcommand{\arraystretch}{0.95}
  \begin{adjustbox}{width=\textwidth, max totalheight=0.92\textheight}
  \begin{tabular}{llccccc}
  \toprule
  \multicolumn{2}{c}{\multirow{4}{*}{\textbf{Benchmark}}} & \multicolumn{5}{c}{\textbf{Model}} \\
  \cmidrule(lr){3-7}
  & & \multirow{2}{*}{\textbf{\modelname}} & \textbf{GLM-4.6V} & \textbf{Qwen3-VL} & \textbf{InternVL} & \textbf{MiMo-VL} \\
  & & & \textbf{Flash} & \textbf{Thinking} & \textbf{3.5} & \textbf{RL-2508} \\
  & & \textit{10B} & \textit{9B} & \textit{8B} & \textit{8B} & \textit{7B} \\
  \midrule[0.8pt]
  \multirow{13}{*}{\parbox{1.6cm}{\centering\textbf{STEM / Multimodal Reasoning}}}
    &  MMMU & \textbf{78.11}& 71.17& \underline{73.53}& 71.69 & 71.14 \\
    &  MMMU-Pro & \textbf{64.08}& 59.93& \underline{60.94}& 59.11 & 60.29 \\
    &  MathVision & \textbf{70.81}& 54.05& 59.60& 52.05 & \underline{59.65} \\
    &  MathVista & \textbf{83.97}& \underline{82.85}& 78.50& 76.78 & 79.86 \\
    &  LogicVista & \textbf{66.89}& 60.29& \underline{64.37}& 54.03 & 63.37 \\
    &  DynaMath & \textbf{56.39}& 48.40& 45.11 & 39.47 & \underline{51.65} \\
    &  ZeroBench (main) & \textbf{1.00}& 0.50& 0.50 & \underline{0.75}& 0.50 \\
    &  ZeroBench (sub) & \textbf{27.54}& \underline{24.03}& 20.58 & 18.56 & 21.18 \\
    &  MathVerse (vision) & \textbf{75.73}& 71.41& \underline{73.19}& 65.18 & \underline{73.19} \\
    &  We-Math & \textbf{73.03}& 61.86& \underline{67.31}& 51.26 & 63.24 \\
    & VisuLogic & \textbf{29.68}& 26.50& \underline{27.82}& 27.20 & 24.52 \\
    &  PhyX & \textbf{59.45}& 52.28& \underline{57.67}& 50.51 & 56.00 \\
    &  HLE & \textbf{10.73}& 3.82& \underline{5.98}& 4.51 & 5.90 \\
  \midrule
  \multirow{9}{*}{\parbox{1.6cm}{\centering\textbf{Recognition / General VQA}}}
    & MMBench EN & \textbf{92.05}& \underline{91.04}& 90.55& 88.20 & 89.91 \\
    & MMBench CN & \textbf{91.55}& 89.56& \underline{89.75}& 86.24 & 88.79 \\
    &  SimpleVQA & \textbf{53.08}& \underline{52.09}& 48.69& 41.43 & 49.65 \\
    &  MMStar & \textbf{77.48}& \underline{74.26}& 73.58& 69.83 & 72.93 \\
    & HallusionBench & \textbf{64.91}& 59.92& 62.23& 58.79 & \underline{63.53} \\
    &  MMVP & \textbf{68.16}& 63.33& 57.17 & 51.33 & \underline{63.50} \\
    & ReMI & \textbf{67.29}& 60.75& 57.17& 52.65 & \underline{63.13} \\
    & M3GIA & \textbf{78.36}& 76.66 & \underline{76.86}& 62.38 & 65.84 \\
    & DoYouSeeMe & \textbf{67.48}& \underline{65.52}& 56.90& 38.22 & 60.92 \\
  \midrule
  \multirow{3}{*}{\parbox{1.5cm}{\centering\textbf{Counting}}}
    &  CountBench & 88.75& \textbf{90.22}& \underline{88.85}& 82.18 & 83.60 \\
    &  CountQA & \underline{33.69}& \textbf{33.79}& 23.35 & 22.32 & 27.41 \\
    &  PixMo-Count & \underline{70.85}& \textbf{76.42}& 69.51 & 63.24 & 69.98 \\
  \midrule
  \multirow{3}{*}{\parbox{1.5cm}{\centering\textbf{Instruction Following}}}
  & MM-MT-Bench & \underline{8.14}& 6.94& \textbf{8.16}& 7.46 & 8.08\\
    &  MIA-Bench & \underline{92.00}& 89.99& \textbf{92.35}& 90.89 & 90.91 \\
    &  MM-IFEval & \textbf{61.87}& 60.78& 60.93 & 60.80 & \underline{61.28} \\
  \midrule
  \multirow{3}{*}{\parbox{1.5cm}{\centering\textbf{Code}}}
    &  HumanEval-V & \textbf{66.05}& 29.26& 26.94 & 24.31 & \underline{31.96} \\
  & Design2Code & \underline{79.55}& 25.93 & 72.21 & 64.15 & \textbf{82.54} \\
  & Design2Code (hard) & \underline{54.69}& 19.37 & 48.75 & 46.25 & \textbf{59.38} \\
  \midrule
  \multirow{3}{*}{\parbox{1.5cm}{\centering\textbf{OCR}}}
  & OCRBench & \textbf{86.75}& \underline{85.97}& 82.85& 83.70 & 85.40 \\
  & OmniOCR & \underline{76.98}& \textbf{80.24}& 75.53 & 70.97 & 74.38 \\
  & CC-OCR (Multi-Lang-OCR) & \textbf{76.59}& 70.85& 69.25& 66.42 & \underline{72.06}\\
  \midrule
  \multirow{12}{*}{\parbox{1.6cm}{\centering\textbf{2D / 3D Spatial Understanding}}}
    &  BLINK & \textbf{66.79}& \underline{64.90}& 62.78& 55.40& 62.57 \\
    &  CVBench & 83.49& \textbf{86.01}& \underline{84.81}& 77.52& 82.04 \\
    &  MMSI-Bench & \textbf{32.18}& \underline{31.13}& 29.05 & 28.12& 29.60 \\
    &  ERQA & \textbf{48.87}& \underline{45.13}& 44.31& 38.88& 41.94 \\
    &  OmniSpatial & \textbf{51.58}& \underline{49.41}& 46.56 & 47.49& 46.74 \\
    & All-Angles-Bench & \textbf{57.21}& \underline{53.24}& 45.88& 45.29& 51.62 \\
    &  MindCube-tiny & \textbf{62.81}& \underline{45.00}& 41.06 & 34.65& 39.06 \\
    & RealWorldQA & \underline{74.44}& \textbf{76.93}& 71.93& 66.93& 72.78 \\
    & SpatialViz-Bench & \textbf{45.51}& 33.79& \underline{35.15}& 32.42& 28.94 \\
    & STARE & \textbf{61.75}& \underline{56.45}& 54.95& 48.20& 55.06 \\
    & CoreCognition & \textbf{66.69}& 65.11 & 64.04 & 61.70& \underline{65.30} \\
    &  V* & 82.85& \textbf{83.51}& 81.02 & 66.89& \underline{83.38} \\
    &  ViewSpatial & \textbf{46.14}& 43.26 & \underline{45.20}& 35.96 & 40.19 \\
  \midrule
  \multirow{4}{*}{\parbox{1.6cm}{\centering\textbf{Document \& Chart Understanding}}}
    & CharXiv (RQ) & \underline{59.52}& \textbf{59.70}& 53.48& 47.15 & 59.32 \\
    & AI2D & \textbf{89.35}& \underline{88.93}& 83.32& 82.34 & 84.96 \\
    &  CSVQA & \textbf{63.33}& 54.99& \underline{61.32}& 46.62 & 60.23 \\
    & OmniDocBench\textsubscript{\tiny NED$\downarrow$} & \textbf{21.51} & 24.88& 23.38 & 29.47& \underline{23.30}\\
  \midrule
  \multirow{4}{*}{\parbox{1.6cm}{\centering\textbf{GUI Grounding}}}
    &  ScreenSpot-Pro & \textbf{51.55}& 45.68& \underline{46.60*}& 15.39& 34.84\\
    &  OSWorld-G & \textbf{59.02}& 54.71& \underline{56.70*}& 31.91& 50.54\\
    & ScreenSpot-V2 & \underline{92.61}& 92.14& \textbf{93.60*}& 84.02& 90.82\\
    & MMBench-GUI (L2) & \textbf{81.50}& \underline{78.46}& 76.60 & 63.95& 75.69\\
  \bottomrule
  \end{tabular}
  \end{adjustbox}
  \end{table}

\paragraph{STEM and Multimodal Reasoning.}
\modelname~consistently outperforms competitive open-source models in the 7B–10B regime across all benchmarks targeting mathematical and scientific reasoning. \modelname~achieves \textbf{78.11\%/64.08\%} on MMMU (Standard/Pro) and notably, on MathVision, it surpasses strong baselines such as MiMo-VL-RL-2508 and Qwen3-VL by more than 10 points. These gains can be primarily attributed to sufficient pre-training and scaled RL compute.
\paragraph{Recognition and General VQA.} 
\modelname~consistently exhibits superior performance in visual recognition and VQA tasks, outperforming existing baselines across all evaluated benchmarks. Notably, \modelname~achieves \textbf{92.05\%/91.55\%} on MMBench (EN/CN), establishing the strongest performance among models within the 10B parameter scale. We attribute this exceptional performance to large-scale, high-quality multimodal pre-training, particularly the scaling of the 1.8B Perception Encoder.



\paragraph{2D / 3D Spatial Understanding.} Despite the absence of specific data curation, \modelname~demonstrates remarkable spatial awareness and reasoning capabilities across both 2D and 3D environments. This emergent proficiency underscores its immense potential as a robust baseline for downstream actionable scenarios, such as embodied intelligence and robotic control.


\paragraph{OCR and Document Understanding.}
\modelname~exhibits frontier-class document intelligence, achieving \textbf{86.75\%} on OCRBench and \textbf{89.35\%} on AI2D. This capability stems from our systematic OCR data construction, which combines extensive real-world collection with high-fidelity synthetic generation.

\paragraph{GUI Grounding and Interaction.}
In actionable intelligence, \modelname~dominates the leaderboard with \textbf{92.61\%} on ScreenSpot-V2 and \textbf{59.02\%} on OSWorld-G. These margins validate our \textbf{Trajectory Modeling} approach, where training on granular action trajectories (e.g., CLICK, SCROLL) effectively grounds visual elements into executable actions, surpassing methods relying solely on static captioning. Notably, these gains are further amplified by RL integrated with a perception reward system, which significantly enhances the model's ability to generalize in complex GUI environments.


\subsection{Text-Centric Evaluation Results}

Table~\ref{tab:same_size_bmk_text} illustrates that \modelname~preserves high-fidelity linguistic intelligence while scaling multimodal training. Unlike prior VL models, \modelname~effectively avoids the performance trade-off between text and vision modalities.

\begin{table}[htbp]
    \centering
    \scriptsize
    \caption{Comparison with SOTA open-source models (7B--10B) on text-centric benchmarks.}
    \label{tab:same_size_bmk_text}
    \setlength{\tabcolsep}{8pt}
    \renewcommand{\arraystretch}{0.95}
    \begin{adjustbox}{max width=\textwidth, max totalheight=0.92\textheight}
    \begin{tabular}{llccccc}
    \toprule
    \multicolumn{2}{c}{\multirow{4}{*}{\textbf{Benchmark}}} & \multicolumn{5}{c}{\textbf{Model}} \\
    \cmidrule(lr){3-7}
    & & \multirow{2}{*}{\textbf{\modelname}} & \textbf{GLM-4.6V} & \textbf{Qwen3-VL} & \textbf{InternVL} & \textbf{MiMo-VL} \\
    & & & \textbf{Flash} & \textbf{Thinking} & \textbf{3.5} & \textbf{RL-2508} \\
    & & \textit{10B} & \textit{9B} & \textit{8B} & \textit{8B} & \textit{7B} \\
    \midrule[0.8pt]
    \multirow{5}{*}{\parbox{1.6cm}{\centering\textbf{Exam}}}
    & MMLU-Pro & 76.02 & 72.30 & \textbf{77.09}& \underline{76.03}& 73.81 \\
    & GPQA-Diamond & \textbf{70.83}& 49.37 & \underline{67.55}& 65.12 & 59.97 \\
    & SuperGPQA & \textbf{50.38}& 42.95 & \underline{49.52}& 42.35 & 45.69 \\
    & LiveBench(2024-11-25) & \underline{69.71}& 44.11 & \textbf{70.79}& 55.62 & 54.35 \\
    \midrule
    \multirow{6}{*}{\parbox{1.6cm}{\centering\textbf{Mathematics}}}
    & AIME2024 & \textbf{90.94}& 37.92 & 74.06& \underline{78.18} & 75.36 \\
    & AIME2025 & \textbf{87.66}& 33.02 & 62.92& 62.50 & \underline{66.51} \\
    & HMMT25 & \textbf{78.18}& 19.17 & 45.21& 35.78 & \underline{47.34} \\
    & CNMO2024 & \underline{78.20}& 61.72 & \textbf{79.22}& 66.56 & 76.17 \\
    & BeyondAIME & \underline{63.23}& 11.80 & 30.59 & \textbf{66.56}& 43.94 \\
    & IMO-AnswerBench & \textbf{62.12}& 22.62 & 38.69 & 35.00 & \underline{48.44} \\
    \midrule
    \multirow{1}{*}{\parbox{1.6cm}{\centering\textbf{Code}}}
    & LiveCodeBench (2408-2505) & \textbf{75.77}& 22.17 & \underline{51.05}& 45.90 & 39.65 \\
    \midrule
    \multirow{3}{*}{\parbox{1.6cm}{\centering\textbf{Instruction Following}}}
    & IFEval & \underline{82.16}& 74.86 & \textbf{83.23}& 79.72 & 68.62 \\
    & IFBench & \textbf{43.28}& 27.47 & \underline{36.65}& 28.14 & 23.47 \\
    & MultiChallenge & \textbf{62.64}& 42.49& \underline{49.82}& 37.73 & 44.69 \\
    \midrule
    \multirow{2}{*}{\parbox{1.6cm}{\centering\textbf{Subjective}}}
    & Arena-Hard-V2 & \textbf{58.57}& 9.26 & \underline{47.34}& 15.57 & 28.59 \\
    & WildBench & \textbf{86.04}& 34.04& \underline{72.36}& 56.45 & 63.09 \\
    \midrule
    \multirow{1}{*}{\parbox{1.6cm}{\centering\textbf{Medical}}}
    & HealthBench & \underline{44.67}& 31.80 & \textbf{47.45}& 35.54 & 43.58 \\
    \bottomrule
    \end{tabular}
    \end{adjustbox}
    \end{table}

\paragraph{Mathematics and Code.} \modelname~significantly outpaces its counterparts in complex reasoning tasks. Its exceptional performance on challenging benchmarks like IMO-AnswerBench (62.12\%) and LiveCodeBench (2408-2505) (75.77 \%) serves as a strong testament to its robust logical inference capabilities, positioning it as a leading model for tasks requiring high-level problem-solving skills.

\paragraph{Human Alignment.} The exceptional instruction-following capabilities and subjective performance of \modelname~further reveal its superior alignment with human preferences, effectively bridging the usability gap traditionally associated with models of 10B size. Our internal Elo-based evaluation confirms that, \modelname~achieves a preference score that matches significantly larger open-source models, demonstrating its potential for high-quality, real-world deployment.



\subsection{Comparison with Larger Models}
\begin{table}[htbp]
  \centering
  \scriptsize
  \caption{Comparisons against models that are 10$\times$–20$\times$ larger, as well as leading proprietary systems, on multimodal and text-centric benchmarks. \textbf{SeRe} and \textbf{PaCoRe} refer to \textbf{Sequential Reasoning} and \textbf{Parallel Coordinated Reasoning}~\citep{hu2025pacore}, respectively.
  }
  \label{tab:larger_size_bmk}
  \setlength{\tabcolsep}{3pt}
  \renewcommand{\arraystretch}{0.90}
  \begin{adjustbox}{width=\textwidth, max totalheight=0.95\textheight}
  \begin{tabular}{llcccccc}
  \toprule
  \multicolumn{2}{c}{\multirow{5}{*}{\textbf{Benchmark}}} & \multicolumn{6}{c}{\textbf{Model}} \\
  \cmidrule(lr){3-8}
  & & \multicolumn{2}{c}{\textbf{\modelname}} & \multirow{2}{*}{\textbf{GLM-4.6V}} & \textbf{Qwen3-VL} & \textbf{Gemini-2.5} & \textbf{Seed-1.5-VL} \\
  \cmidrule(lr){3-4}
  & & \textbf{SeRe} & \textbf{PaCoRe} & & \textbf{Thinking} & \textbf{Pro} & \textbf{Thinking} \\
  & & \textit{10B} & \textit{10B} & \textit{106B-A12B}& \textit{235B-A22B} & \textit{—} & \textit{—} \\
  \midrule[0.8pt]
  
  
  \multicolumn{8}{c}{\textbf{Multimodal Benchmarks}} \\
  \midrule[0.5pt]
  
  \multirow{13}{*}{\parbox{1.5cm}{\centering\textbf{STEM / Multimodal Reasoning}}}
    &  MMMU & 78.11& \underline{80.11}& 75.20 & 78.70& \textbf{83.89}& 79.11\\
    &  MMMU-Pro & 64.08& 67.18 & 65.84 & \underline{72.37}& \textbf{76.96}& 70.60\\
    &  MathVision & 70.81& \textbf{75.95}& 63.50* & 72.10& \underline{73.30*}& 68.70*\\
    &  MathVista & 83.97& \underline{85.50}& 83.51 & 85.10& 83.88& \textbf{85.60}\\
    &  LogicVista & 66.89& 71.36 & 64.88 & \textbf{73.15}& 69.80& \underline{72.93} \\
    &  DynaMath & 56.39& \textbf{61.48}& 56.29 & \underline{60.30}& 52.30& 58.88 \\
    &  ZeroBench (main) & 1.00& \textbf{5.00}& 1.00 & 3.00& \underline{4.00}& 1.00 \\
    &  ZeroBench (sub) & 27.54& 29.94 & 29.04 & 28.40& \textbf{33.53}& \underline{31.74} \\
    &  MathVerse (vision)& 75.73& \textbf{78.30}& 72.84 & 76.65 & \underline{78.30}& 77.79 \\
    &  We-Math & 73.03& 73.90 & 71.14 & 74.70 & \textbf{80.10}& \underline{79.05} \\
    & VisuLogic & 29.68& \underline{32.70}& 28.30 & 31.80& 31.40 & \textbf{34.30} \\
    &  PhyX & 59.45& 66.01 & 59.70 & \underline{66.30}& \textbf{67.56}& 62.53 \\
  \midrule
  \multirow{9}{*}{\parbox{1.5cm}{\centering\textbf{Recognition / General VQA}}}
    & MMBench EN & 92.05& 92.38 & \underline{92.75}& 92.70& \textbf{93.19}& 92.11 \\
    & MMBench CN & 91.55& \underline{91.96}& 91.88 & 91.80 & \textbf{93.13}& 91.76 \\
    &  SimpleVQA & 53.08& 54.64 & 57.95 & 59.30& \textbf{66.85}& \underline{64.72} \\
    &  MMStar & 77.48& 77.64 & 75.30 & 76.80& \textbf{79.18}& \underline{77.91} \\
    & HallusionBench & 64.91& 64.54 & 60.63 & \underline{65.58}& \textbf{65.63}& 64.13 \\
    &  MMVP & 68.16& 68.00 & \underline{71.33}& 71.30 & 70.67& \textbf{74.00} \\
    & ReMI & 67.29& 69.12 & 64.42 & \textbf{74.70}& 71.69 & \underline{72.19} \\
    & M3GIA & 78.33& 73.50 & 78.72 & 81.00& \underline{83.11}& \textbf{83.22} \\
    & DoYouSeeMe & 67.48& 68.54 & 67.50& \textbf{72.89}& 71.19 & \underline{71.94} \\
  \midrule
  \multirow{3}{*}{\parbox{1.2cm}{\centering\textbf{Counting}}}
    &  CountBench & 88.75& 88.80 & \underline{92.06}& \textbf{92.46}& 87.78& 91.85 \\
    &  CountQA & 33.69& 38.29 & 36.32 & \underline{45.62}& 38.02 & \textbf{48.89} \\
    &  PixMo-Count & 70.85& 71.61 & 76.47 & \underline{79.80}& 75.54 & \textbf{83.38} \\
  \midrule
  \multirow{3}{*}{\parbox{1cm}{\centering\textbf{OCR}}}
    & OCRBench & 86.75& \textbf{89.00}& 86.20 & \underline{87.30}& 85.90 & 85.20 \\
    &  OmniOCR & 76.98& 78.14 & 84.53 & \underline{87.20}& 66.05 & \textbf{87.80} \\
    & CC-OCR (Multi-Lang-OCR) & 76.59& 77.51 & 74.08 & \underline{80.80}& \textbf{81.10}& 78.82 \\
  \midrule
  \multirow{12}{*}{\parbox{1.5cm}{\centering\textbf{2D / 3D Spatial Understanding}}}
    &  BLINK & 66.79& 67.39 & 68.17 & 67.12& \textbf{72.01}& \underline{71.54} \\
    &  CVBench & 83.49& 85.92 & 83.72 & \textbf{87.86}& 84.36& \underline{86.27} \\
    &  MMSI-Bench & 32.18& \underline{36.40}& 30.80 & 32.50 & \textbf{40.40}& 30.60 \\
    &  ERQA & 48.87& 51.75 & 47.75 & \underline{53.50}& \textbf{62.25}& 48.50 \\
    &  OmniSpatial & 51.58& 52.58 & 50.49 & \underline{53.10}& \textbf{55.64}& 51.99 \\
    & All-Angles-Bench & 57.21& \underline{64.71}& 62.94 & 60.59 & \textbf{65.88}& 57.65 \\
    &  MindCube-tiny & \underline{62.81}& \textbf{68.58}& 52.83 & 47.58 & 58.92 & 39.83 \\
    & RealWorldQA & 74.44& 75.56 & 77.78 & \underline{78.80}& 77.78 & \textbf{79.61} \\
    & SpatialViz-Bench & 45.51& \textbf{52.03}& 37.46 & \underline{46.36}& 45.34 & 35.25 \\
    & STARE & 61.75& \underline{64.57}& 60.38 & \textbf{70.89}& 62.36 & 62.99 \\
    & CoreCognition & 66.69& 71.54 & 69.50 & \underline{72.66}& \textbf{78.78}& 72.38 \\
    &  V* & 82.85& 84.29 & 85.86& \underline{89.53}& 80.63 & \textbf{90.58} \\
    &  ViewSpatial & 46.14& \underline{48.41}& 43.87& \textbf{48.58}& 44.15 & 44.14 \\
  
  \midrule[0.8pt] 
  
  
  \multicolumn{8}{c}{\textbf{Text-Centric Benchmarks}} \\
  \midrule[0.5pt]
  
  \multirow{5}{*}{\parbox{1.5cm}{\centering\textbf{Exam}}}
    & MMLU-Pro & 76.02 & 77.09 & 79.96 & \underline{83.75}& \textbf{86.45}& 83.39 \\
    & GPQA-Diamond & 70.83 & 73.99 & 69.19 & \underline{77.68}& \textbf{84.06}& 71.91\\
    & SuperGPQA & 50.38 & 53.15 & 53.28 & \underline{64.20}& \textbf{65.00}& 60.50 \\
    & LiveBench(2024-11-25) & 69.71 & 71.69 & 62.75 & \textbf{80.14}& \underline{76.34}& 65.62\\
  \midrule
  \multirow{6}{*}{\parbox{1.5cm}{\centering\textbf{Mathematics}}}
    & AIME2024 & 90.94 & \textbf{93.33}& 80.63 & \underline{91.93}& 79.53& 79.48\\
    & AIME2025 & \underline{87.66}& \textbf{94.43}& 71.88& 83.59& 83.96& 64.06\\
    & HMMT25 & \underline{78.18}& \textbf{92.14}& 57.29& 67.71& 65.68& 51.30\\
    & CNMO2024 & 78.20 & 81.17 & 72.11 & \textbf{88.36}& 74.53& \underline{83.67}\\
    & BeyondAIME & \underline{63.23}& \textbf{74.00}& 39.83 & 57.42& 54.45& 42.83\\
    & IMO-AnswerBench & 62.12 & \textbf{76.66}& 51.25 & 69.25& \underline{72.00}& 44.75\\
  \midrule
  \multirow{1}{*}{\parbox{1.5cm}{\centering\textbf{Code}}}
    & LiveCodeBench (2408-2505) & \underline{75.77}& \textbf{76.43}& 48.71 & 69.45& 72.01& 57.10\\
  \bottomrule
  \end{tabular}
  \end{adjustbox}
\end{table}
To evaluate the performance ceiling of \modelname, we benchmark it against strong open-source (10$\times$--20$\times$ larger) and closed-source flagships. As shown in Table~\ref{tab:larger_size_bmk}, \modelname~effectively bridges the gap between limited parameter scales (10B) and frontier intelligence. On standard benchmarks, \modelname~outperforms GLM-4.6V (106B-A12B) across perception, recognition, and complex reasoning tasks, while remaining competitive with Qwen3-VL-Thinking (235B-A22B). Notably, it achieves 70.81\% on MathVision, 87.66\% on AIME 2025, and 77.48\% on MMStar, demonstrating exceptional multimodal intelligence within a compact 10B budget.

We further explore the model's limits by scaling test-time compute via the parallel coordinated reasoning setting. As shown in Table~\ref{tab:larger_size_bmk}, the \textbf{PaCoRe} mode of \modelname~consistently surpasses its standard SeRe mode and achieves frontier-level performance on several reasoning-heavy and perception-centric benchmarks, even outperforming Gemini-2.5-Pro and Seed-1.5-VL.
Specifically, \modelname~achieves 80.11\% on the multimodal understanding and reasoning benchmark MMMU. On the challenging multimodal mathematical reasoning benchmarks, MathVision and MathVista, it scores 75.95\% and 85.50\%, respectively. Furthermore, on representative visual recognition tasks such as MMBench and MMStar, it attains 92.17\% (average on CN \& EN) and 77.64\%, respectively. These results demonstrate that \modelname~has reached a leading level in multimodal perception and reasoning.
Even more notably, on challenging high-level textual mathematics tasks like AIME2025 and HMMT25, it achieves remarkable scores of 94.43\% and 92.14\%, respectively. These results significantly outperform competing models, underscoring that intelligence is not strictly constrained by model size.

\section{Discussion}
\label{sec:discuss}


This section presents a two-fold analysis of the empirical findings that shaped \modelname. First, we distill key design insights regarding model architecture and optimization strategies that informed our final configurations. Second, we characterize the learning dynamics during RLVR and the emergent capabilities arising from subsequent RL scaling.

\subsection{Ablations and Design Insights}
\label{ablation}


\paragraph{Vision Encoder Selection: PE-lang vs. DINOv3.}
We compare the Perception Encoder (PE-lang, 300M parameters specifically selected for ablation) with DINOv3 (ViT-large-16, 300M parameters)~\citep{siméoni2025dinov3} as the vision backbone. 
While DINOv3 excels in pure vision tasks, it suffers from slow convergence in our multimodal setting due to the modality gap. Conversely, PE-lang explicitly pre-aligned with LLMs achieves superior data efficiency and benchmark performance (\Cref{tab:vision_encoder_comp}). This underscores that \textbf{language alignment in the vision encoder remains a prerequisite for efficient VL modeling}, irrespective of subsequent trillion-scale generative training.

\begin{table}[h]
\centering
\caption{Comparison of Vision Encoders: \textbf{DINOv3} vs. \textbf{PE-lang}. Here, \textbf{Omni.} and \textbf{SVQA} denote \textbf{OmniSpatial} and \textbf{SimpleVQA}, respectively.}
\label{tab:vision_encoder_comp}
\resizebox{\linewidth}{!}{
\begin{tabular}{p{4cm}|cccc|cccccc}
\toprule
\textbf{Vision Encoder} & \multicolumn{4}{c|}{\textbf{Perception}} & \multicolumn{6}{c}{\textbf{General}} \\
 & \textbf{BLINK} & \textbf{Omni.} & \textbf{MMVP} & \textbf{OCRBench} & \textbf{MMStar} & \textbf{SVQA} & \textbf{CCBench} & \textbf{V*} & \textbf{MMMU} & \textbf{ReMI} \\
\midrule
DINOv3 & \textbf{42.35} & 43.31 & 28.00 & 57.60 & 41.43 & \textbf{22.18} & 56.32 & 34.55 & 46.56 & 24.50 \\
\textbf{PE-lang (Ours)} & 41.19 & \textbf{43.57} & \textbf{32.00} & \textbf{70.10} & \textbf{42.10} & 21.15 & \textbf{59.39} & \textbf{37.17} & \textbf{47.67} & \textbf{26.08} \\
\rowcolor[gray]{0.95} \textit{$\Delta$} & \textit{-1.16} & \textit{+0.26} & \textit{+4.00} & \textit{+12.50} & \textit{+0.67} & \textit{-1.03} & \textit{+3.07} & \textit{+2.62} & \textit{+1.11} & \textit{+1.58} \\
\bottomrule
\end{tabular}
}
\vspace{2pt}
\end{table}

\paragraph{Optimizer Choice: Muon vs. AdamW.}
We investigate Muon~\citep{muon}, a matrix-wise optimizer utilizing Newton-Schulz iteration~\citep{bernstein2024oldoptimizernewnorm} to regularize weight topology. Muon effectively addresses the noise and imbalance inherent in large-scale multimodal data, yielding notable improvements in \Cref{tab:muon_vs_adamw} for 
tail-knowledge tasks (+6.48\% SimpleVQA). These results suggest Muon effectively \textbf{reduces sensitivity to data scarcity}. Despite these capabilities, \textbf{we exclude Muon from the final architecture due to initialization mismatch}. Recent literature~\citep{liu2025muonscalablellmtraining} indicates that Muon is sensitive to weights initially optimized by element-wise methods like AdamW. In our setting, this necessitates a prolonged warmup period to stabilize the transition, which paradoxically limits overall training efficiency compared to a well-tuned AdamW baseline. We therefore leave a more thorough exploration of Muon for future work.

\begin{table}[htbp]
\centering
\caption{\textbf{AdamW vs. Muon} optimizers across selected benchmarks.}
\label{tab:muon_vs_adamw}
\resizebox{\linewidth}{!}{
\begin{tabular}{p{4cm}|cccc|cccccc}
\toprule
\textbf{Optimizer} & \multicolumn{4}{c|}{\textbf{Perception}} & \multicolumn{6}{c}{\textbf{General}} \\
 & \textbf{BLINK} & \textbf{Omni.} & \textbf{MMVP} & \textbf{OCRBench} & \textbf{MMStar} & \textbf{SVQA} & \textbf{CCB} & \textbf{V*} & \textbf{MMMU} & \textbf{ReMI} \\
\midrule
Muon & \textbf{41.14} & 42.73 & \textbf{32.00} & 67.70 & \textbf{44.58} & \textbf{27.08} & \textbf{60.72} & 36.65 & \textbf{47.56} & 22.23 \\
\textbf{Adam (Ours)} & 40.72 & \textbf{44.94} & 29.33 & \textbf{71.10} & 41.77 & 20.60 & 60.13 & \textbf{39.27} & 46.11 & \textbf{25.00} \\
\rowcolor[gray]{0.95} \textit{$\Delta$} & \textit{-0.42} & \textit{+2.21} & \textit{-2.67} & \textit{+3.40} & \textit{-2.81} & \textit{-6.48} & \textit{-0.59} & \textit{+2.62} & \textit{-1.45} & \textit{+2.77} \\
\bottomrule
\end{tabular}
}
\end{table}



\paragraph{Ablation for Deepstack.}

We investigate the utility of Deepstack~\citep{meng2024deepstack}, a depth-extension technique successfully utilized in Qwen3-VL~\citep{bai2025qwen3vltechnicalreport}. 
While enabling Deepstack effectively accelerates training convergence, this optimization-level improvement does not translate into meaningful gains on downstream evaluation benchmarks, as shown in ~\Cref{tab:deepstack_ablation}. Given the computational overhead versus the marginal utility, we exclude it from the final model configuration.

\begin{table}[htbp]
\centering
\caption{Ablation study of \textbf{Deepstack} architecture scaling.}
\label{tab:deepstack_ablation}
\resizebox{\linewidth}{!}{
\begin{tabular}{p{4cm}|cccc|cccccc}
\toprule
\textbf{Technique} & \multicolumn{4}{c|}{\textbf{Perception}} & \multicolumn{6}{c}{\textbf{General}} \\
 & \textbf{BLINK} & \textbf{Omni.} & \textbf{MMVP} & \textbf{OCRBench} & \textbf{MMStar} & \textbf{SVQA} & \textbf{CCB} & \textbf{V*} & \textbf{MMMU} & \textbf{ReMI} \\
\midrule
w/ DeepStack & \textbf{40.72} & 42.92 & 26.00 & \textbf{71.20} & \textbf{43.31} & \textbf{28.66} & \textbf{63.94} & 36.65 & 47.44 & \textbf{26.96} \\
\textbf{w/o DeepStack (Ours)} & 40.61 & \textbf{43.57} & \textbf{31.33} & 69.30 & 42.44 & 25.20 & 62.80 & \textbf{38.22} & \textbf{47.78} & \textbf{26.96} \\
\rowcolor[gray]{0.95} \textit{$\Delta$} & \textit{-0.11} & \textit{+0.65} & \textit{+5.33} & \textit{-1.90} & \textit{-0.87} & \textit{-3.46} & \textit{-1.14} & \textit{+1.57} & \textit{+0.34} & \textit{+0.00} \\
\bottomrule
\end{tabular}
}
\end{table}



\subsection{RL Dynamics, Performance, and Emergence}
\label{exp_rl}

\begin{figure}[t]
    \centering
    \includegraphics[width=\linewidth]{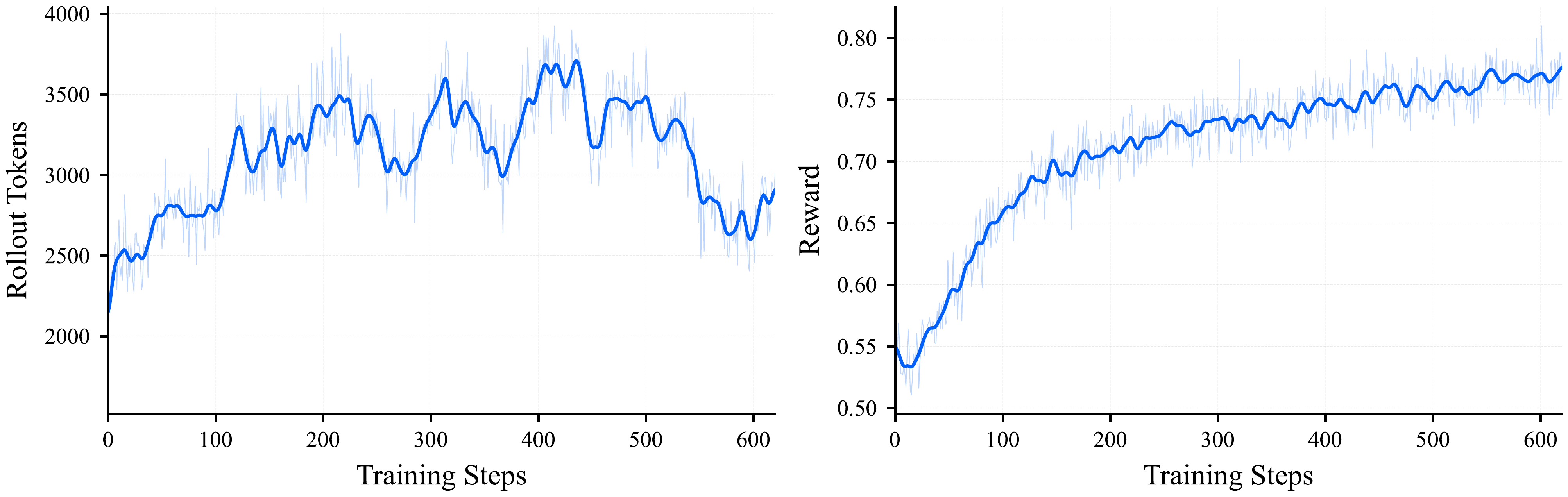}
    \caption{\textbf{RLVR dynamics.} While the reward continuously increases without saturating (right), the average rollout tokens decrease towards the starting level after an initial rise (left).}
    \label{fig:rl_curve}
\end{figure}

\begin{figure}[t]
    \centering
    \includegraphics[width=\linewidth]{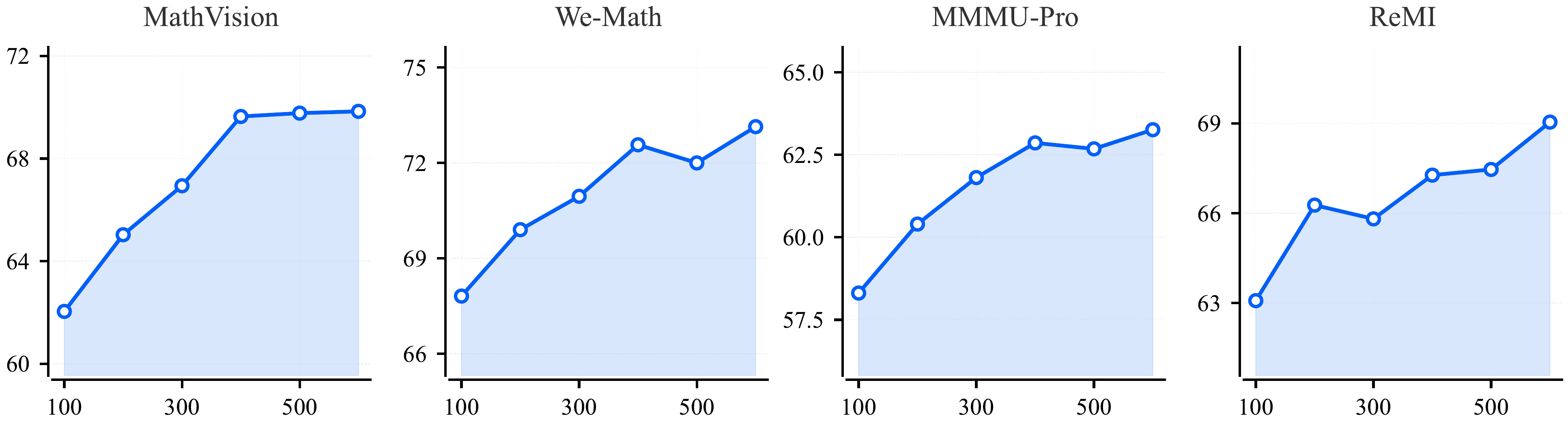}
    \caption{\textbf{Trends of representative multimodal reasoning and perception metrics during RLVR.} Evaluated every 100 iterations, performance mirrors the reward dynamics: rapid initial growth followed by steady improvement.}
    \label{fig:rl_bmk_trend}
\end{figure}

\paragraph{Training Dynamics and Continuous Improvement.} We track the evolution of RLVR over 600 training iterations, monitoring reward progression, average rollout length, and downstream performance across multimodal reasoning, recognition, OCR, and grounding tasks (assessed every 100 iterations). As illustrated in \Cref{fig:rl_curve} (right) and \Cref{fig:rl_bmk_trend}, the model exhibits a robust two-phase growth trajectory: an initial rapid ascent in both rewards and metrics during the first 200 iterations, followed by a steady, linear increase. Remarkably, the reward consistently approaches $0.8$ \textbf{without observing saturation}, mirrored by continuous gains in downstream metrics.

\paragraph{Distinct Length Dynamics.} In contrast to the ``sequential scaling'' (i.e., the progressive lengthening of reasoning paths) typically observed in text-only RL~\citep{dsr1_cite,hu2025open}, the \textbf{average rollout length} in \modelname~does not increase monotonically. Instead, it rises initially but eventually returns to its starting level (see \Cref{fig:rl_curve}, left). We identify this as a \textbf{cancellation effect} between two opposing scaling properties:

\begin{enumerate} 
    \item \textbf{Reasoning Tasks (e.g., STEM, Puzzles)}: Exhibit \textbf{standard sequential scaling}, where model performance is positively correlated with the extension of inference-time compute (i.e., chain-of-thought length).
    \item \textbf{Deterministic Perception Tasks (e.g., Grounding, OCR)}: Characterized by \textbf{length diminishment} via policy refinement. Unlike the expansive ``thinking'' chains required for reasoning, RL gains in perception stem from \textbf{entropy reduction}~\citep{cui2025entropy}. To be specific, RL optimization induces a systematic collapse of the search space by pruning redundant exploratory tokens. This process concentrates the probability mass onto the singular deterministic mode, effectively converting high-temperature \textbf{Pass@N} exploration into robust \textbf{Pass@1} accuracy~\citep{passn}. In this regime, shorter rollout lengths serve as a direct proxy for higher model confidence and sharpened perceptual focus.
\end{enumerate}

\paragraph{The "Missing Trace" Hypothesis.} Given that RL scaling strictly depends on the initial policy's behavioral coverage~\citep{wang2025octothinker,fan2025megascience}, we hypothesize that the perceptual "unscaling" stems from a critical data deficiency. While human visual cognition relies on iterative, coarse-to-fine processes (e.g., "glance-and-focus" or "try-error-correct" mechanisms akin to o3~\citep{openaio3_cite} or RePer~\citep{wei2025perception}), these internal mental states are rarely explicitly verbalized in training corpora. Consequently, the RL optimization landscape lacks the "cognitive traces" necessary to spontaneously incentivize sequential perceptual reasoning.

\begin{table}[h]
    \centering
    \begin{tabular}{@{}p{0.95\textwidth}@{}}
        \toprule
        \textbf{Question:} Locate the mish and Mennonite section in the image. <image> 
        $\dots$ \\
        Now, based on the original problem and reference responses above, please provide your own comprehensive solution. \\
        
        \midrule
        
        \textbf{Response:} <think> \\
        Got it, let's look at the references. \\
        \textbf{The references have different x-coordinates (like 220, 228, 227, etc.) but the common y is 185 or around there.} \\
        Let's check the problem's context. \\
        The references' labels mention $\dots$ \\
        Let's see the reference 5 has 228, reference 6 227, reference 9 228, etc. \\
        To create a new point, we need to follow the logic. Let's check the references' labels and their coordinates' variations. \\
        $\dots$ \\
        Maybe the key is to find a value that's in the middle of the range. Alternatively, check the problem's structure.
        $\dots$ \\
        \bottomrule
    \end{tabular}
    \vspace{0.5em}
    
    \caption{A perception case of parallel coordinated reasoning: The model first aggregates reference answers, then cross-validates their validity, and ultimately identifies the underlying patterns of the final answer.}
    \label{tabs:emergence}
\end{table}

\paragraph{Emergence via PaCoRe.} To address this limitation, we employ PaCoRe~\citep{hu2025pacore}, a test-time scaling strategy that effectively explicates these implicit visual processes. Conceptually, this approach mirrors the \textit{proposal-then-refinement} philosophy of foundational computer vision architectures, such as the Region Proposal Networks (RPN) in Faster R-CNN~\citep{ren2016fasterrcnnrealtimeobject}. In this framework, the model first samples multiple perceptual hypotheses (proposals); conditioning on these, the model then performs a multi-faceted synthesis to derive the final output. As illustrated in \Cref{tabs:emergence}, this process spontaneously gives rise to complex, \textbf{human-like verification behaviors} during inference.

Crucially, this paradigm exhibits two distinctive scaling properties: (1) A steady, deliberate \textbf{growth in response length}, indicating the model's ability to effectively allocate additional compute for hypothesis verification. (2) \textbf{Significant performance gains} in PaCoRe mode over the vanilla SeRe mode, as shown in \Cref{tab:larger_size_bmk}. These gains are evident across benchmarks demanding intensive reasoning, such as MathVision (+5.14\%) and DynaMath (+5.09\%), as well as those requiring \textbf{exhaustive perception} (especially depends on high recall rate), including visual counting (CountQA, +4.6\%), OCR (OCRBench, +2.25\%), and especially, spatial understanding (All-Angles-Bench, +7.50\%, SpatialViz-Bench, +6.52\%).

\paragraph{Compress System 2 to System 1.} 
PaCoRe, functioning as a primitive multi-agent framework, indeed enables perceptual scaling, where the proposer generates massive visual proposals in parallel, and the controller subsequently performs sequential cross-checking and self-verification. Looking forward, we aim to employ self-distillation to internalize these materialized, parallel coordinated reasoning traces. By injecting the logic of parallel deliberation directly into the model’s parameters, we seek to transform expensive ``slow-thinking''~\citep{kahneman2011thinking} traces into high-fidelity, intrinsic intuition, ultimately fostering a more efficient and accurate perceptual foundation.

\section{Conclusion and Future Work}
\label{sec:limitations}

Anchored by a rigorously curated corpus of 1.2T multimodal tokens and sharpened via over 1k iterations of sequential and parallel coordinated RL, \modelname~has achieved capabilities in perception, reasoning, and alignment that rival the strongest proprietary and open-source frontiers. Yet, raw capability is not synonymous with systemic maturity. On the trajectory toward comprehensive multimodal intelligence, we identify critical bottlenecks in computational density and physical grounding. Our strategic roadmap aims to transform these limitations into the next engines of growth:

\paragraph{Maximizing Token Efficiency via Universal RL Scaling.} 
We prioritize the principle that every unit of compute, during both training and inference, must contribute directly to intelligence density.

\begin{itemize} 
    \item \textbf{Shifting Compute from Pre-training to RL.} RL scaling demonstrates continuous, saturation-free performance leaps that pre-training alone cannot sustain. We intend to aggressively pivot computational resources toward RL. By scaling universally in both \textbf{depth} (sequential reasoning) and \textbf{width} (parallel exploration), we aim to uncover high-value perception and reasoning traces, pushing the upper bounds of multimodal intelligence for models of all scales. 
    \item \textbf{Optimizing Reasoning Density.} We aim to bridge the gap between the high performance of extensive search and the low latency of standard inference. Our goal is to \textbf{internalize} the benefits of parallel exploration and eliminate redundant ``over-thinking.'' We envision a regime that continuously compresses reasoning paths, transforming explicit, coordinated search into efficient sequentiality, and ultimately distilling these capabilities into instinctive, ``System 1''-like responses.
\end{itemize}

\paragraph{Bridging the Reality Gap.} While the model excels in digital tasks, the "reality gap" remains the critical frontier. We posit that bridging this gap necessitates a paradigm shift: moving beyond passive data consumption to active physical grounding.

\begin{itemize} 
    \item \textbf{From Semantic to Physical World Models.} We regard current text-based multi-agent synthesis as a foundational step—constructing a semantic world model. To achieve true embodiment, we must scale this synthesis to encompass massive video trajectories and sensorimotor action sequences. This unifies distinct modalities into a \textbf{holistic world model} that transcends linguistic logic to internalize physical causality and spatiotemporal dynamics. 
    \item \textbf{Physics as the Ultimate Verifier.} Current multimodal RL often relies on static or noisy proxy labels. We intend to integrate high-fidelity simulation environments where rewards are strictly governed by immutable physical laws. This shifts the learning paradigm from \textbf{surface-level imitation to interaction-driven mastery}, grounding the model's reasoning in verifiable causality rather than statistical correlation. 
    \item \textbf{Embodied Chain-of-Thought (E-CoT).} We envision extending the reasoning context to explicitly model temporal dynamics and physical state transitions. By training the model to articulate ``physical intuition'' via predicting dynamics prior to action, we aim to develop agents capable of robust long-horizon planning in dynamic, open-world environments.
\end{itemize}

\newpage
\bibliography{main}

\newpage
\section{Author List}

All authors are listed in alphabetical order by their first names.
$^{\dagger}$ indicates project leaders.

\paragraph{Core Contributors}

Ailin Huang, Chengyuan Yao, Chunrui Han, Fanqi Wan, Hangyu Guo, Haoran Lv, Hongyu Zhou, Jia Wang, Jian Zhou, Jianjian Sun$^{\dagger}$, Jingcheng Hu, Kangheng Lin, Liang Zhao$^{\dagger}$, Mitt Huang, Song Yuan, Wenwen Qu, Xiangfeng Wang, Yanlin Lai, Yingxiu Zhao, Yinmin Zhang, Yukang Shi, Yuyang Chen, Zejia Weng, Ziyang Meng

\paragraph{Contributors} 

Ang Li, Aobo Kong, Bo Dong, Changyi Wan, David Wang, Di Qi, Dingming Li, En Yu, Guopeng Li, Haiquan Yin, Han Zhou, Hanshan Zhang, Haolong Yan, Hebin Zhou, Hongbo Peng, Jiaran Zhang, Jiashu Lv, Jiayi Fu, Jie Cheng, Jie Zhou, Jisheng Yin, Jingjing Xie, Jingwei Wu, Jun Zhang, Junfeng Liu, Kaijun Tan, Kaiwen Yan, Liangyu Chen, Lina Chen, Mingliang Li, Qian Zhao, Quan Sun, Shaoliang Pang, Shengjie Fan, Shijie Shang, Siyuan Zhang, Tianhao You, Wei Ji, Wuxun Xie, Xiaobo Yang, Xiaojie Hou, Xiaoran Jiao, Xiaoxiao Ren, Xiangwen Kong, Xin Huang, Xin Wu, Xing Chen, Xinran Wang, Xuelin Zhang, Yana Wei, Yang Li, Yanming Xu, Yeqing Shen, Yuang Peng, Yue Peng, Yu Zhou, Yusheng Li, Yuxiang Yang, Yuyang Zhang, Zhe Xie, Zhewei Huang, Zhenyi Lu, Zhimin Fan, Zihui Cheng

\paragraph{Sponsors} 

Daxin Jiang, Qi Han, Xiangyu Zhang, Yibo Zhu, Zheng Ge

\newpage
\appendix
\section*{Appendix}

\section{More Results}
\label{sec:cases_appendix}
\textbf{Qualitative Results.}
This section presents a selection of qualitative examples illustrating the capabilities of \textbf{\modelname} through responses generated from various prompts. The provided cases underscore the model’s superior ability of fine-grained perception and structured logical reasoning. More importantly, it demonstrates a high degree of cognitive systematicity, effectively integrating domain knowledge to execute multi-step tasks.
\\
\\
\large \textbf{Case 1: Visual Pattern Matching and Sequential Decoding}

\begin{figure}[h!]
\centering
\includegraphics[width=0.50\textwidth]{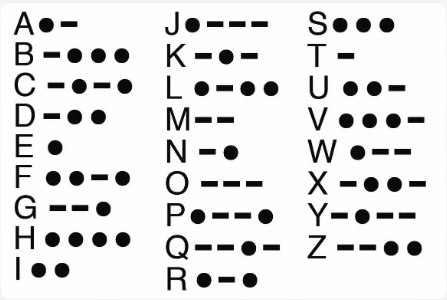}
\caption{Morse Code Reference}
\label{fig:case1_morse}
\end{figure}

\begin{caseprompt}
\begin{CJK*}{UTF8}{gbsn}
根据该表生成"STEPFUN"的编码
\end{CJK*}
\end{caseprompt}

\begin{casethinking}
{\small
\begin{CJK*}{UTF8}{gbsn}
用户现在需要把每个字母转换成莫尔斯电码，先看每个字母在表中的编码。首先分解"STEPFUN"每个字母：S、T、E、P、F、U、N。先逐个找：\\
\textbf{$\dots$}
\begin{itemize}[leftmargin=*]
    \item S：看右边第三列第一个，S是$\bullet\bullet\bullet$（三个点）
    \item T：第三列第二个，T是$-$（一个短横）
    \item E：左边第一列第五个，E是$\bullet$（一个点）
    \item P：中间列（J到R）中P对应的符号是$\bullet--\bullet$（点、短横、短横、点）
    \item F：最左侧列第六行，F的符号是$\bullet\bullet-\bullet$（两个点、一个短横、一个点）
    \item U：最右侧列第三行，U的符号是$\bullet\bullet-$（两个点、一个短横）
    \item N：中间列（J到R）第五行，N的符号是$-\bullet$（一个短横、一个点）
\end{itemize}
\textbf{$\dots$} \\
再将每个字母的编码按顺序拼接（字母间用空格分隔更清晰），"STEPFUN"的莫尔斯电码为：
$\bullet\bullet\bullet \quad - \quad \bullet \quad \bullet--\bullet \quad \bullet\bullet-\bullet \quad \bullet\bullet- \quad -\bullet$
\end{CJK*}
}
\end{casethinking}

\begin{caseanswer}
$\bullet\bullet\bullet \quad - \quad \bullet \quad \bullet--\bullet \quad \bullet\bullet-\bullet \quad \bullet\bullet- \quad -\bullet$
\end{caseanswer}

\newpage
\large \textbf{Case 2: Fine-grained Visual Counting and Semantic Filtering}

\begin{figure}[h!]
\centering
\includegraphics[width=0.65\textwidth]{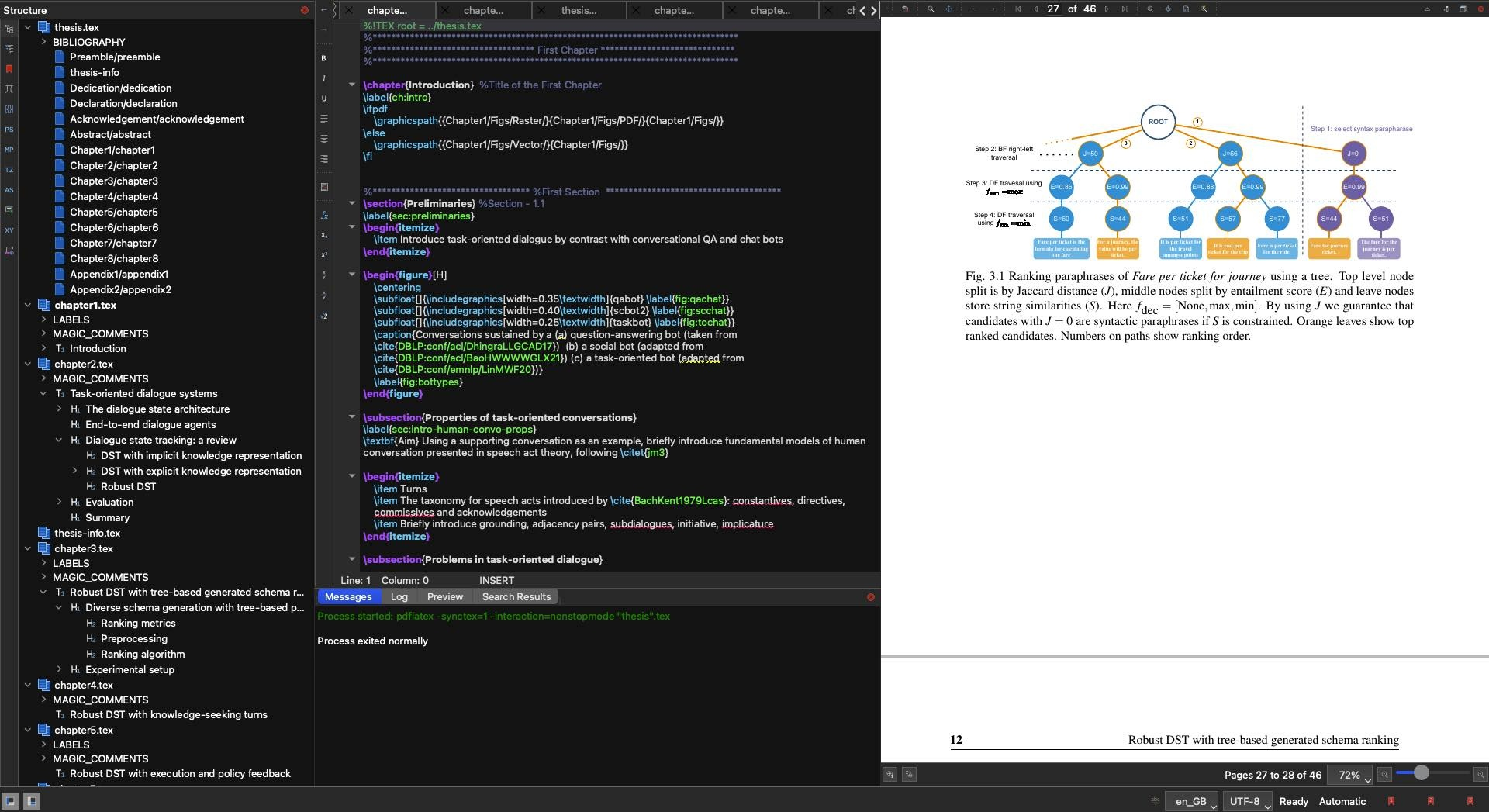}
\caption{Screenshot of the Compiler}
\label{fig:case2_code}
\end{figure}

\begin{caseprompt}
The screenshot shows a snapshot of a LaTeX text editor a student uses for writing their PhD dissertation. What is the minimum number of chapter .tex files the student has opened?
\end{caseprompt}

\begin{casethinking}
So, let's look at the tabs at the top of the editor. \\
\textbf{$\dots$}\\
The visible tabs are:
\begin{itemize}[leftmargin=*]
    \item chapte$\dots$ (leftmost, maybe chapter1.tex as the active file in the center)
    \item chapte$\dots$ (next)
    \item thesis$\dots$ (this is thesis.tex, not a chapter)
    \item chapte$\dots$ (next)
    \item chapte$\dots$ (next)
    \item chapte$\dots$ (last one before the dark theme part)
\end{itemize}
\textbf{$\dots$}\\
To determine the minimum number of chapter .tex files opened, we analyze the tabs at the top of the editor.\\
1. Identify tabs: The image shows multiple tabs labeled chapte$\dots$ (truncated for "chapter$\dots$") and one tab labeled thesis$\dots$.  \\
2. Exclude non-chapter tabs: The thesis$\dots$ tab is not a chapter file, so we count only the chapte$\dots$ tabs. \\ 
3. Count chapter tabs: There are 5 visible chapte$\dots$ tabs (excluding the thesis$\dots$ tab).  
\end{casethinking}

\begin{caseanswer}
5
\end{caseanswer}

\newpage
\large \textbf{Case 3: Low-Resolution Visual Detail Recognition}

\begin{figure}[h!]
\centering
\includegraphics[width=0.55\textwidth]{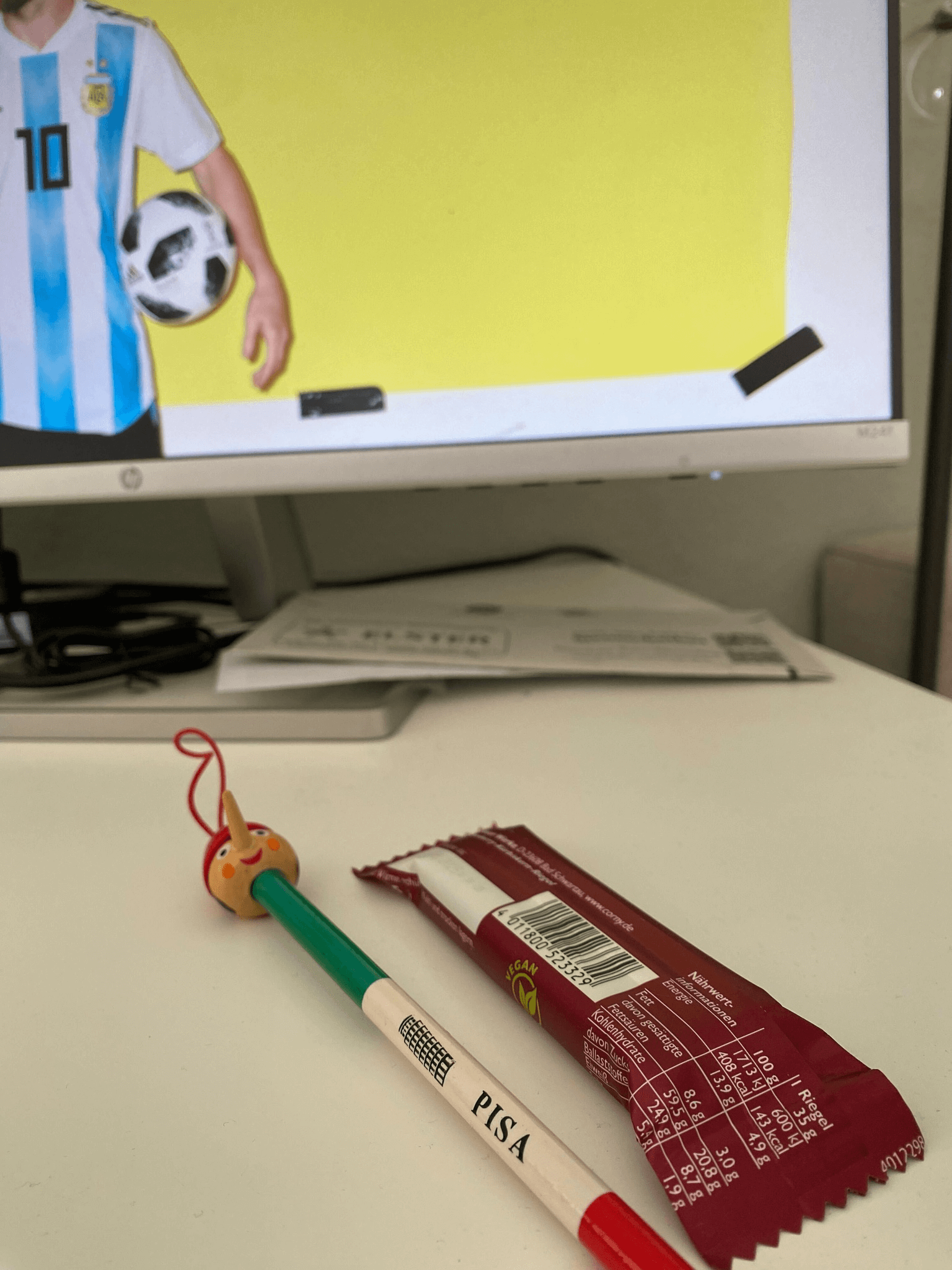}
\caption{Out-of-focus Target}
\label{fig:case3_monitor}
\end{figure}

\begin{caseprompt}
What is the brand name of the monitor in the image? What is its encoding with a Caesar cipher with a left shift of 5. Give your answer as a combined string.
\end{caseprompt}

\begin{casethinking}
Got it, let's break this down. First, find the monitor's brand. \\
\textbf{$\dots$} \\
The logo on the monitor's base is HP. \\
Then, we need to do a Caesar cipher with a left shift of 5. \\
\textbf{$\dots$}
\begin{itemize}[leftmargin=*]
    \item First, H is the 8th letter (A=1, B=2$\dots$ H=8). Left shift 5: 8-5=3 → C.
    \item Then, P is the 16th letter. 16-5=11 → K.
\end{itemize}
So "HP" shifted left 5 is "CK". 
\end{casethinking}

\begin{caseanswer}
HPCK
\end{caseanswer}

\newpage
\large \textbf{Case 4: Topological Graph Analysis and Minimum Weight Search}

\begin{figure}[h!]
\centering
\includegraphics[width=0.75\textwidth]{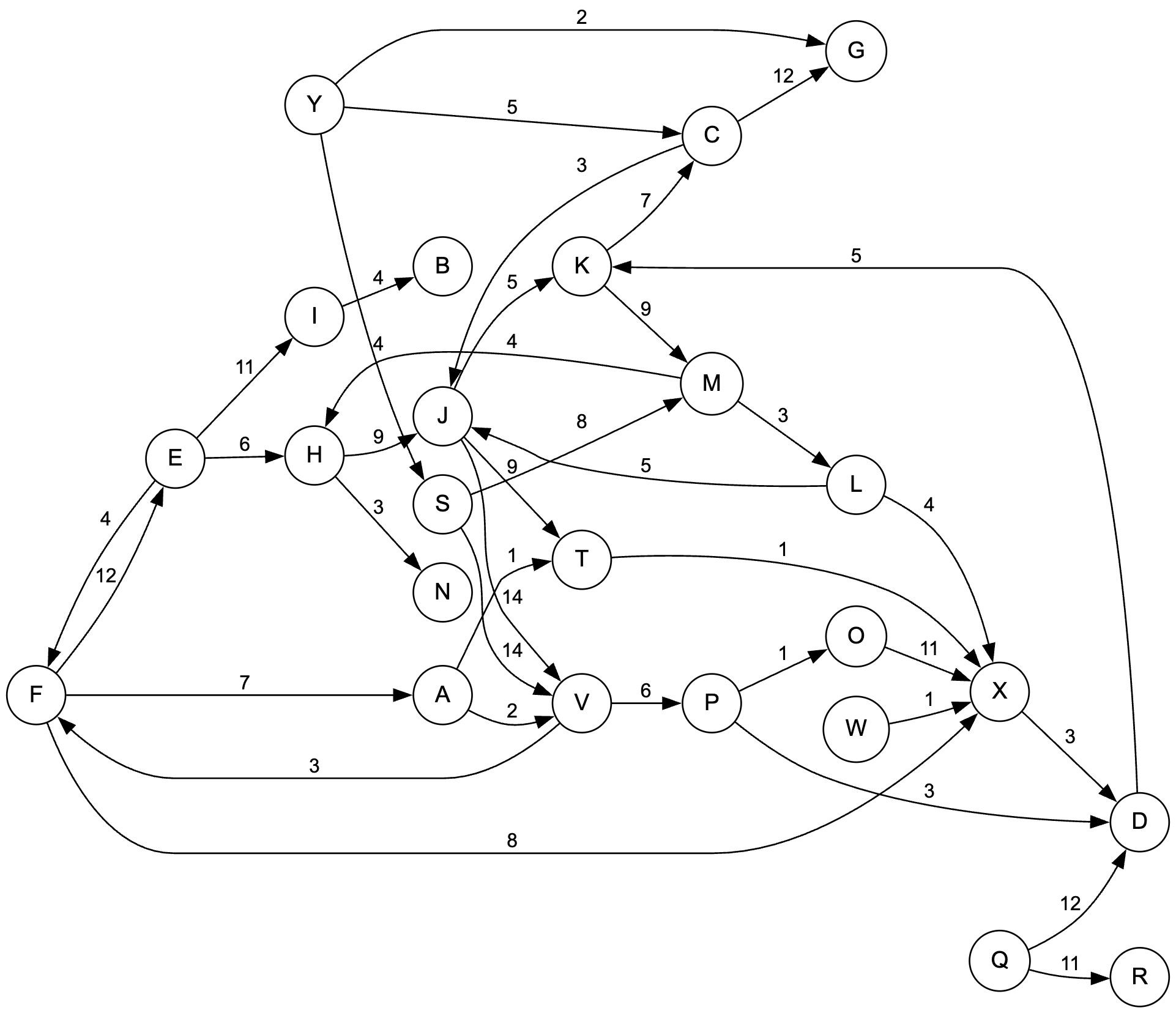}
\caption{Weighted Graph}
\label{fig:case4_graph}
\end{figure}

\begin{caseprompt}
How many edges have the shortest length?
\end{caseprompt}

\begin{casethinking}
To determine the number of edges with the shortest length, we first identify the smallest edge weight in the diagram. \\
\textbf{$\dots$}\\
The smallest weight is 1. Next, we count all edges labeled with 1:\\
\textbf{$\dots$}
\begin{itemize}[leftmargin=*]
    \item $A \rightarrow T$: labeled 1
    \item $T \rightarrow X$: labeled 1
    \item $P \rightarrow O$: labeled 1
    \item $W \rightarrow X$: labeled 1
\end{itemize}
\textbf{$\dots$}\\
No other edges have a weight of 1. Thus, there are 4 edges with the shortest length.

\end{casethinking}

\begin{caseanswer}
4
\end{caseanswer}

\section{Serialization Details for Synthesis in PaCoRe}
\label{sec:synthesis}

\begin{table}[h]
    \centering
    \begin{tcolorbox}[colback=white, colframe=black, arc=0pt, outer arc=0pt, boxrule=0.5pt, left=10pt, right=10pt, top=10pt, bottom=10pt]
        \small
        \noindent You are given a problem and a list of reference responses. Your job is to analyze these references and provide your own response. \\
        Original Problem: \\
        \{\{ original\_prompt \}\} \\
        
        \noindent Reference Responses: \\
        \{\% for response in ref\_responses \%\} \\
        Reference \{\{ loop.index \}\}: \\
        \{\{ response \}\} \\
        \{\% endfor \%\} \\
        
        \noindent Now, based on the original problem and reference responses above, please provide your own comprehensive solution.
    \end{tcolorbox}
    \vspace{-5pt}
    \caption{\textbf{Input serialization template for PaCoRe synthesis.} We use this template to embed the current problem $x$ (denoted as \texttt{original\_prompt}) and the compact message set $M$ (denoted as \texttt{ref\_responses}) into the model's context. In the degenerate case where the message set is empty ($M = \emptyset$), this template is bypassed, and the original problem input is passed to the model unmodified.}
    \label{tab:serialization_template}
\end{table}

As detailed in Table~\ref{tab:serialization_template}, we frame compact messages as ``Reference Responses'' to encourage the model to synthesize diverse perspectives. By populating the ``Original Problem'' slot with the latest observation while maintaining the interaction history in context, PaCoRe ensures seamless compatibility with existing reasoning ecosystems. Further implementation details regarding the synthesis process can be found in \citet{hu2025pacore}, Section C.

\section{Evaluation Details}
\label{sec:data_appendix}
This section outlines the evaluation setup and the corresponding evaluation prompts.


\subsection{Evaluation Details for Multimodal Benchmarks}
\label{sec:evaluation_prompts}
We detail the prompt formats used for evaluation across different benchmarks. For each benchmark, we present the corresponding prompt template, where \{question\} denotes the textual problem description, potentially including answer options, and <image> represents the visual input. When images are embedded in the question with explicit positional semantics, their original positions are preserved; otherwise, images are placed before the question text.

\paragraph{MMMU.} 
We adopt the evaluation metric suggested by OpenCompass.\footnote{https://github.com/open-compass/VLMEvalKit}. The placement of image placeholders follows the original MMMU samples, allowing for interleaved visual inputs.

\promptbox{
<image>\\
\{question\}\\
Your response can be freely expressed in any format, but the final answer must be presented in this format:\\
"Final answer: [the correct option]" with the option letter only.
}

\paragraph{MMMU-Pro.} We use the official metric of MMMU-Pro.

\promptbox{
<image>\\
\{question\}\\
Your response can be freely expressed in any format, but the final answer must be presented in this format:\\
"Final answer: [the correct option]" with the option letter only.
}

\paragraph{MathVision.} We use the official metric of MathVision.

\promptbox{
Please solve the problem and put your answer in one "\textbackslash boxed\{\}". If it is a multiple choice question, only one letter is allowed in the "\textbackslash boxed\{\}". \\
<image>\\
\{question\}
}

\paragraph{MathVista.} 
For MathVista, we follow the official evaluation protocol and use distinct prompt templates corresponding to different answer formats.\footnote{https://github.com/lupantech/MathVista}

\noindent For questions requiring floating-point answers with one or two decimal places, we use the following prompts, respectively:

\promptbox{
<image>\\
Hint: Please answer the question requiring a floating-point number with two decimal places and provide the final value, e.g., 1.23, 1.34, 1.45, at the end.\\
Question: \{question\}
}

\promptbox{
<image>\\
Hint: Please answer the question requiring a floating-point number with one decimal place and provide the final value (e.g., 1.2, 1.3, 1.4) at the end.\\
Question: \{question\}
}

\noindent For multiple-choice questions, we use the following prompt:

\promptbox{
<image>\\
Hint: Please answer the question and provide the correct option letter (e.g., A, B, C, D) at the end.\\
Question: \{question\}
}

\noindent For questions requiring an integer answer, we use the following prompt:

\promptbox{
<image>\\
Hint: Please answer the question requiring an integer answer and provide the final value (e.g., 1, 2, 3) at the end.\\
Question: \{question\}
}

\noindent For questions requiring a Python list as the answer, we use the following prompt:

\promptbox{
<image>\\
Hint: Please answer the question requiring a Python list as an answer and provide the final list, e.g., [1, 2, 3], [1.2, 1.3, 1.4], at the end.\\
Question: \{question\}
}

\noindent Additional details are available on the official MathVista website.

\paragraph{LogicVista.} We use the official metric of LogicVista.

\promptbox{
<image>\\
\{question\}\\
Your response can be freely expressed in any format, but the final answer must be presented in this format:\\
"Final answer: [the correct option]" with the option letter only.
}

\paragraph{DynaMath.} For DynaMath, we adopt the official evaluation protocol and use the worst-case accuracy metric, defined as the percentage of correctly answered seed questions across all generated variations, to assess model robustness on mathematical reasoning tasks.\footnote{https://github.com/DynaMath/DynaMath}

\noindent For multiple-choice questions, we use the following prompt:

\promptbox{
<image>\\
\{question\}\\
Your response can be freely expressed in any format, but the final answer must be presented in this format:\\
"Final answer: [the correct option]" (option letter only).
}

\noindent For questions requiring a floating-point answer, we use the following prompt:

\promptbox{
<image>\\
\{question\}\\
Your response can be freely expressed in any format, but the final answer must be presented in this format:\\
"Final answer: \textbackslash boxed\{\{answer\}\}." Round the answer to three decimal places.
}

\noindent For all other questions, we use the following prompt:

\promptbox{
<image>\\
\{question\}\\
Your response can be freely expressed in any format, but the final answer must be presented in this format:\\
"Final answer: \textbackslash boxed\{\{answer\}\}."
}

\noindent Additional details are available on the official DynaMath repository.

\paragraph{ZeroBench.} We use the official metric of ZeroBench.

\promptbox{
<image>\\
\{question\}\\
Give the final answer in curly braces, like: \textbackslash boxed\{final\_answer\}.
}

\paragraph{MathVerse.} We use the official metric of MathVerse and focus on the Vision-only subset. Details of the answer extraction and judgement can be seen in the official MathVerse repository. \footnote{https://github.com/ZrrSkywalker/MathVerse}

\noindent For multiple-choice questions, we use the following prompt:

\promptbox{
Answer the question in the image. Provide the correct option letter, e.g., A, B, C, D, within \textbackslash boxed\{\}. \\
<image>
}

\noindent For other questions, we use the following prompt:

\promptbox{
Answer the question in the image. Put your answer within \textbackslash boxed\{\}. \\
<image>
}

\paragraph{We-Math.} We use the official metric of We-Math. 

\promptbox{
<image>\\
\{question\} \\
Your response can be freely expressed in any format, but the final answer must be presented in this format:\\
"Final answer: [the correct option]" with the option letter only.
}

\paragraph{VisuLogic.} We use the official metric of VisuLogic.

\promptbox{
<image>\\
\{question\} \\
Your response can be freely expressed in any format, but the final answer should follow this format:\\
Answer: \textbackslash boxed\{\$LETTER\}.
}

\paragraph{PhyX.} We use the official metric of PhyX.

\promptbox{
<image>\\
\{question\} \\
Your response can be freely expressed in any format, but the final answer must be presented in this format:\\
"Final answer: [the correct option]" with the option letter only.
}

\paragraph{HLE.} We follow the official evaluation metrics and LLM-based judgement protocols of HLE.

\promptbox{
<image>\\
\{question\}
}

\paragraph{MMBench.} We report accuracy on the MMBench v1.1 Dev set. We use the following prompt for MMBench-EN, and apply its Chinese translation for MMBench-CN.

\promptbox{
<image>\\
\{question\} \\
Your response can be freely expressed in any format, but the final answer must be presented in this format:\\
"Final answer: [the correct option]" with the option letter only.
}

\paragraph{SimpleVQA.}  We follow the official evaluation metrics and LLM-based judgement protocols of SimpleVQA.

\promptbox{
<image>\\
\{question\}
}

\paragraph{MMStar.} We use official metric of MMStar.

\promptbox{
<image>\\
\{question\} \\
Your response can be freely expressed in any format, but the final answer must be presented in this format:\\
"Final answer: [the correct option]" with the option letter only.
}

\paragraph{HallusionBench.} We use official metric of HallusionBench.

\promptbox{
<image>\\
\{question\} \\
Please answer yes or no.
}

\paragraph{MMVP.} We use the official metric of MMVP. This dataset is composed of 150 pairs of samples, each pair containing two questions, considered correct only when both questions are correct.

\promptbox{
<image>\\
\{question\} \\
Please select the correct answer from the options above. \\
Your response can be freely expressed in any format, but the final answer must be presented in this format:\\
"Final answer: [the correct option]" with the option letter only.
}

\paragraph{ReMI.} We use the official metric of ReMI. The placement of
image placeholders follows the original ReMI samples, allowing for interleaved visual inputs.

\promptbox{
<image>\\
\{question\} \\
Please select the correct answer from the options above. \\
Your response can be freely expressed in any format, but the final answer must be presented in this format:\\
"Final answer: [the correct option]" with the option letter only.
}

\paragraph{M3GIA.} We use the official metric and adopt the following system and user prompts.

\noindent We use the system prompt format below:

\promptbox{
 Answer following questions with the option's letter from the given choices directly.
}

\paragraph{DoYouSeeMe.} We use the official metric and adopt the following system and user prompts.

\noindent We adopt the system prompt specified by the question domain.

\promptbox{
\# ['Shape Discrimination', 'Joint Shape-Color', 'Spatial Grids']\\
You are an AI assistant specialized in visual perception tasks. Please analyze the image carefully and provide your answer as an integer number. Format your response by putting your final answer after 'Answer:', for example: Answer: 5 \\ \\
\# ['Letter Discrimination'] \\
You are an AI assistant specialized in visual perception tasks. Please analyze the image carefully and identify the letter or text. Format your response by putting your final answer after 'Answer:', for example: Answer: A \\ \\
\# ['Form Constancy', 'Visual Closure', 'Visual Figure-Ground']
You are an AI assistant specialized in visual perception tasks. Please analyze the image carefully and select the correct option. Format your response by putting your final answer after 'Answer:', using only the option number (1-4), for example: Answer: 2 \\ \\
\# Otherwise:\\
You are an AI assistant specialized in visual perception tasks. Please analyze the image carefully and provide your answer. Format your response by putting your final answer after 'Answer:'
}

\noindent And we use the user prompt format below:

\promptbox{
<image>\\
\{question\}
}


\paragraph{CountBench.} We use the official metric of CountBench.

\promptbox{
<image>\\
\{question\} \\
Please select the correct answer from the options above. \\
Your response can be freely expressed in any format, but the final answer must be presented in this format:\\
"Final answer: [the correct count number]" with the number only.
}

\paragraph{CountQA.} We use the official metric of CountQA.

\promptbox{
<image>\\
\{question\} \\
Please select the correct answer from the options above. \\
Your response can be freely expressed in any format, but the final answer must be presented in this format:\\
"Final answer: [the correct answer]".
}

\paragraph{PixMo-Count.} We use the official metric of PixMo-Count.

\promptbox{
<image>\\
\{question\} \\
Please select the correct answer from the options above. \\
Your response can be freely expressed in any format, but the final answer must be presented in this format:\\
"Final answer: [the correct count number]" with the number only.
}

\paragraph{MM-MT-Bench.} We use the official metric of MM-MT-Bench.\footnote{https://github.com/mistralai/mistral-evals/tree/main}

\promptbox{
<image>\\
\{question\}
}

\paragraph{MIA-Bench.} We follow the official evaluation metrics and LLM-based judgement protocols of MIA-Bench.

\promptbox{
<image>\\
\{question\}
}

\paragraph{MM-IFEval.} We follow the official metric and evaluation protocols. We adopt the system prompts specified by the question types.

\noindent For the P-Level questions, we use the following system prompt:

\promptbox{
    You are an AI assistant. Please answer the question based on the image. Provide a clear and concise answer.
}

\noindent For the C-Level questions, we use the following system prompt:

\promptbox{
    \# Have constraints \\
    You are an AI assistant. Please answer the question based on the image 
    while strictly following these constraints: \\
    \{constrains\} \\
    Make sure your response adheres to ALL the constraints above.
    
    \# Others \\
    You are an AI assistant. Please answer the question based on the image 
    while following any instructions or constraints mentioned in the question.
}

\noindent And we use the user prompt format below:

\promptbox{
<image>\\
\{question\}
}

\paragraph{HumanEval-V.} We use the official metric of HumanEval-V.

\promptbox{
<image>\\
\{question\}
}

\paragraph{Design2Code.} We use the official metric of Design2Code.

\promptbox{
You are an expert web developer who specializes in HTML and CSS.
A user will provide you with a screenshot of a webpage. \\
You need to return a single html file that uses HTML and CSS to reproduce the given website. \\
Include all CSS code in the HTML file itself. \\
If it involves any images, use "rick.jpg" as the placeholder.\\
Some images on the webpage are replaced with a blue rectangle as the placeholder, use "rick.jpg" for those as well.\\
Do not hallucinate any dependencies to external files. You do not need to include JavaScript scripts for dynamic interactions.\\
Pay attention to things like size, text, position, and color of all the elements, as well as the overall layout.\\
Respond with the content of the HTML+CSS file:\\
<image>
}

\paragraph{OCRBench 
.} We use the official metric and follow the prompt.

\promptbox{
<image>\\
\{question\}
}

\paragraph{Omni-OCR.} We use the official metric of Omni-OCR.

\promptbox{
<image>\\
\{question\}
}

\paragraph{CC-OCR (Multi-Lang-OCR subset).} We use the official metric of CC-OCR.

\promptbox{
<image>\\
Please output only the text content from the image without any additional descriptions or formatting.
}

\paragraph{BLINK.} We use the official metric of BLINK.

\promptbox{
<image>\\
\{question\} \\
Your response can be freely expressed in any format, but the final answer must be presented in this format:\\
"Final answer: [the correct option]" with the option letter only.
}

\paragraph{CVBench.} We use the official metric of CVBench.

\promptbox{
<image>\\
\{question\}
}

\paragraph{MMSI-Bench.} We use the official metric of MMSI-Bench.

\promptbox{
<image>\\
\{question\}\\
Answer with the option's letter from the given choices directly. Enclose the option's letter within ` '.
}

\paragraph{ERQA.} We use the official metric of ERQA.

\promptbox{
<image>\\
\{question\} \\
Your response can be freely expressed in any format, but the final answer must be presented in this format:\\
"Final answer: [the correct option]" with the option letter only.
}

\paragraph{OmniSpatial.} We use the official metric of OmniSpatial.

\promptbox{
<image>\\
\{question\} \\
Your response can be freely expressed in any format, but the final answer must be presented in this format:\\
"Final answer: [the correct option]" with the option letter only.
}

\paragraph{All-Angles-Bench.} We use the official metric of All-Angles-Bench.

\promptbox{
<image>\\
\{question\} \\
Your response can be freely expressed in any format, but the final answer must be presented in this format:\\
"Final answer: [the correct option]" with the option letter only.
}

\paragraph{MindCube-tiny.} MindCube-tiny is a condensed subset of the MindCube benchmark designed for efficient evaluation of Vision-Language Models in reconstructing 3D spatial structures and performing mental simulations from limited perspectives. We use its official metric.

\promptbox{
<image>\\
\{question\} \\
Your response can be freely expressed in any format, but the final answer must be presented in this format:\\
"Final answer: [the correct option]" with the option letter only.
}

\paragraph{RealWorldQA.} We use the official metric of RealWorldQA.

\promptbox{
<image>\\
\{question\} \\
Your response can be freely expressed in any format, but the final answer must be presented in this format:\\
"Final answer: [the correct option]" with the option letter only.
}

\paragraph{SpatialViz-Bench.}
We use the official metric of SpatialViz-Bench.

\promptbox{
<image>\\
\{question\} \\
You should first provide a reasoning process, then provide a single option(A, B, C or D) as the final answer. The reasoning process and the answer are enclosed within <think></think> and <answer></answer> tags, respectively, i.e., <think>reasoning process</think>, <answer>answer</answer>.

}

\paragraph{STARE.}
We use the official metric of STARE.

\promptbox{
<image>\\
\{question\} \\
Answer with the option's letter from the given choices and put the letter in one \textbackslash boxed\{\}.\\
Please solve the problem step by step.
}

\paragraph{CoreCognition.}
We use the official metric of CoreCognition.

\noindent For multiple-choice questions, we use the following prompt: \\
\promptbox{
<image>\\
\{question\} \\
Answer with the option's letter from the given choices and put the letter in one \textbackslash boxed\{\}.
}

\noindent For the True/False (Yes/No) questions, we use the following prompt: \\
\promptbox{
<image> \\
\{question\} \\
Answer with YES or NO and put the answer in one \textbackslash boxed\{\}.
}

\paragraph{V*.} We use the official metric of V*.

\promptbox{
<image>\\
\{question\} \\
Your response can be freely expressed in any format, but the final answer must be presented in this format:\\
"Final answer: [the correct option]" with the option letter only.
}

\paragraph{ViewSpatial.}
We use the official metric of ViewSpatial.

\promptbox{
<image>\\
\{question\} \\
Reply only to the corresponding option.\\
Answer:
}

\paragraph{CharXiv (RQ).} We use the official metric of CharXiv and adopt the following system and user prompts.

\noindent We adopt the system prompt specified by the question domain.

\promptbox{
You should first think about the reasoning process in the mind and then provide the user with the answer. The reasoning process is enclosed within <think> </think> tags, i.e. <think> reasoning process here </think> answer here.
}

\noindent And we use the user prompt format below:

\promptbox{
<image>\\
\{question\}
}

\paragraph{AI2D.} We use the official metric of AI2D.

\promptbox{
Answer following questions with the option's letter from the given choices directly. \\
<image>\\
\{question\}
}

\paragraph{CSVQA.} We use the official metric of CSVQA.

\noindent For multiple-choice questions, we use the following prompt:

\begin{CJK*}{UTF8}{gbsn}
\promptbox{
<image>\\
请回答图片中的问题。将正确选项字母（如A、B、C、D）放在 \textbackslash boxed\{\} 中。
}
\end{CJK*}

\noindent For other questions, we use the following prompt:

\begin{CJK*}{UTF8}{gbsn}
\promptbox{
<image>\\
请回答图片中的问题。将最终答案放在 \textbackslash boxed\{\} 中。
}
\end{CJK*}


\paragraph{EncQA.} We use the official metric and adopt the following user prompts specified by the question type.

\promptbox{
\# For multiple choices problem \\
Answer using only a single word or letter from the options provided. \\
\{question\} \\
Options: \{options\} \\
\\
\# For set problems \\
Answer choosing only from the options provided, your answer should be just a simple comma separated list. \\
\{question\} \\
Options: \{options\} \\
\\
\# For numeric problem \\
Answer using only a single number. \\
\{question\} \\
}

\paragraph{OmniDocBench.} We use the NED (Normalized Edit Distance) metric and adopt the official system prompt and user prompt. \footnote{https://github.com/opendatalab/OmniDocBench}

\paragraph{ScreenSpot-Pro \& ScreenSpot-V2 \& OSWorld-G \& MMBench-GUI-L2.} For the GUI grounding tasks, we use a unified prompt format designed to localize visual elements and output their coordinates in a structured form.

\promptbox{
<image>\\
Based on the instruction `\{question\}', locate the target element and output its coordinate point in JSON format.
}

\subsection{Evaluation Details for Text-Centric Benchmarks}

To reduce metric variance and improve result reliability on text-centric benchmarks, we perform repeated evaluation for selected benchmarks. For a benchmark with Repeat = $N$, each sample is evaluated independently $N$ times, and the final score is reported as the average over all runs.

The repetition settings for each text-centric benchmark are listed below:

\begin{itemize}
  \item \textbf{MMLU-Pro}: Repeat = 1
  \item \textbf{GPQA-Diamond}: Repeat = 16
  \item \textbf{SuperGPQA}: Repeat = 1
\item \textbf{LiveBench(2024-11-25)}: Repeat = 1
  \item \textbf{AIME 2024}: Repeat = 64
  \item \textbf{AIME 2025}: Repeat = 64
  \item \textbf{HMMT25}: Repeat = 64
  \item \textbf{CNMO2024}: Repeat = 64
  \item \textbf{BeyondAIME}: Repeat = 64
\item \textbf{IMO-AnswerBench}: Repeat = 1
  \item \textbf{LiveCodeBench (2408-2505)}: Repeat = 16
  \item \textbf{IFEval}: Repeat = 4
  \item \textbf{IFBench}: Repeat = 4
  \item \textbf{MultiChallenge}: Repeat = 1
  \item \textbf{Arena-Hard-V2}: Repeat = 1
  \item \textbf{WildBench}: Repeat = 1
  \item \textbf{HealthBench}: Repeat = 1
\end{itemize}

\subsection{Evaluation Details for Ablations}
Each ablation study in \Cref{ablation} is conducted on checkpoints pre-trained with the same number of billions of tokens, ensuring fair and controlled comparisons, but without extending to the final checkpoint due to computational cost.
In terms of evaluation setups, these results are attained from a few-shot evaluation manner on the pre-trained checkpoints.

\setcounter{figure}{0}
\makeatletter
\renewcommand{\thefigure}{A\@arabic\c@figure}
\makeatother

\setcounter{table}{0}
\makeatletter
\renewcommand{\thetable}{A\@arabic\c@table}
\makeatother

\end{document}